\pdfoutput=1
\documentclass[conference,compsoc]{IEEEtran}

\usepackage{amssymb}
\usepackage[nocompress]{cite}
\usepackage{amsmath,amsfonts}
\usepackage{graphicx}
\usepackage{subcaption}
\usepackage{caption}
\usepackage{textcomp}

\usepackage{xcolor}
\usepackage{hyperref}
\hypersetup{
    colorlinks=true,
    linkcolor=blue,
    filecolor=magenta,      
    urlcolor=cyan,
    pdftitle={Overleaf Example},
    pdfpagemode=FullScreen,
    }

\usepackage{microtype}
\usepackage{booktabs} 
\usepackage{multirow}  %
\usepackage{tcolorbox}
\usepackage{diagbox}
\usepackage{amsbsy}
\usepackage{multicol}
\usepackage{lipsum}
\usepackage{pifont}
\usepackage[utf8]{inputenc}
\usepackage[english]{babel}
\usepackage{amsthm}

\usepackage{tikz}
\usepackage{pbox}
\usepackage{color,colortbl,array,xspace}
\usepackage{todonotes}

\newcommand{\para}[1]{\vspace{2pt}\noindent{\textbf{#1}}\hspace{10pt}\vspace{0.1pt}}

\newenvironment{icompact}{
  \begin{list}{$\bullet$}{
    \itemindent -.05em
    \parsep 0pt plus 1pt
    \partopsep 0pt plus 1pt
    \topsep 2pt plus 2pt minus 2pt
    \itemsep 0pt plus 1.3pt
    \parskip 0pt plus 2pt
    \leftmargin 0.13in}
      }
{\normalsize
\end{list}
}

\usepackage{scalerel,graphicx,xparse}
\usepackage{amsmath}
\usepackage{filecontents}
\usepackage{amsthm}
\newcommand{\sys}{{{SneakyPrompt}}\xspace}
\newcommand{\sysb}{{{SneakyPrompt-base}}\xspace}
\newcommand{\sysr}{{{SneakyPrompt-RL}}\xspace}
\usepackage{mathrsfs}
\usepackage{multicol}
\usepackage{mdframed}
\usepackage{pifont}
\usepackage{algpseudocode}
\usepackage{algorithm}
\newtheorem{definition}{Definition}

\begin{document}
\sloppy

\title{\sys: Jailbreaking Text-to-image Generative Models} 

\author{
Yuchen Yang,
Bo Hui,
Haolin Yuan,
Neil Gong$^\dagger$, and
Yinzhi Cao \\
\{yc.yang, bo.hui, hyuan4, yinzhi.cao\}@jhu.edu,  neil.gong@duke.edu\\
Johns Hopkins University, $^\dagger$Duke University
}

\maketitle
\thispagestyle{plain}
\pagestyle{plain}

\begin{abstract}

Text-to-image generative models such as Stable Diffusion and DALL$\cdot$E  raise many ethical concerns due to the generation of harmful images such as Not-Safe-for-Work (NSFW) ones. To address these ethical concerns, safety filters are often adopted to \emph{prevent} the generation of  NSFW images.  In this work, we propose \sys,  the first automated attack framework, to \emph{jailbreak} text-to-image generative models such that they generate NSFW images even if safety filters are adopted. Given a prompt that is blocked by a safety filter, \sys repeatedly queries the text-to-image generative model and strategically perturbs tokens in the prompt based on the query results to bypass the safety filter.   
 Specifically, \sys utilizes reinforcement learning to guide the perturbation of tokens. 
 Our evaluation shows that \sys successfully jailbreaks DALL$\cdot$E 2 with closed-box safety filters to generate  NSFW images.  Moreover, we also deploy several state-of-the-art, open-source safety filters on a Stable Diffusion model.  Our evaluation shows that \sys not only successfully generates NSFW images, but also outperforms existing text adversarial attacks when extended to jailbreak text-to-image generative models, in terms of both the number of queries and qualities of the generated NSFW images.  \sys is open-source and available at this repository: \url{https://github.com/Yuchen413/text2image_safety}.

\end{abstract}
\section{Introduction}

Text-to-image generative models (or called \emph{text-to-image models} for short)---e.g., Stable Diffusion~\cite{Rombach_2022_stablediffusion}, DALL$\cdot$E~\cite{ramesh2022_dalle2}, and Imagen~\cite{saharia2022_imagen}---are popular due to the invention and deployment of diffusion models~\cite{diffusion-1,diffusion-2} and large-scale language models~\cite{CLIP,llm-gpt,llm-t5}.  Such text-to-image models---which generate a synthetic image based on a given text prompt---have broad applications such as graphic design and virtual environment creation.  For example, Microsoft has embedded DALL$\cdot$E~\cite{ramesh2022_dalle2} into an application named Designer~\cite{msdesigner} and an image creator tool as part of Microsoft Edge; in addition, Stable Diffusion has been used by more than 10 million people daily up to February 2023. 
 Yet, one practical ethical concern facing text-to-image models is that they may generate \textit{sensitive} Not-Safe-for-Work (NSFW) images~\cite{NSFW_image, sf_text_match} such as those related to violence and child-inappropriate. Therefore, existing text-to-image models all adopt so-called \textit{safety filters} as guardrails to block the generation of such NSFW images.  However, 
 the robustness of such safety filters---especially those used in practice---to adversarial manipulations of prompts is still unknown. 

 One intuitive method for jailbreaking safety filters is to treat them as closed-boxes and launch text-based adversarial attacks like TextBugger~\cite{TextBugger}, Textfooler~\cite{TextFooler}, BAE~\cite{BAE}, and a concurrent work called adversarial prompting~\cite{maus2023adversarial} to perturb prompts. However, existing text-based attacks focus on misleading a classification model but \textit{not} bypassing safety filters with NSFW generations. For example, 
 none of the aforementioned approaches is able to bypass the closed-box safety filter of  DALL$\cdot$E~2 according to our evaluation. 
   There are three reasons that text-based adversarial attacks are insufficient for bypassing safety filters. 
    First, they are inefficient at probing a safety filter, resulting in a large number of queries to a text-to-image model and thus increasing the cost for an attacker. 
    Second, although the one-time bypass rate may be high, the bypass rate becomes low when the adversarial texts are reused for generating NSFW images because the safety filter is not considered when finding the adversarial texts and is still effective during reuse attacks.   Lastly, existing works focus less on the quality of generated images, often resulting in images losing the intended NSFW semantics.

Two recent works studied the safety filters of text-to-image models. 
 Specifically, Rando et al.~\cite{rando2022_red} reverse engineer Stable Diffusion's safety filter  
  and    then propose a manual bypass strategy that adds 
    extra unrelated text 
    to a prompt. 
    Another concurrent work---Qu et al.~\cite{qu2023unsafe}---manually gathers a template NSFW prompt dataset to evaluate safety filters of open-source text-to-image models, e.g.,  Stable Diffusion.  
    However, the generation of adversarial prompts to bypass safety filters is largely manual, which often results in a low bypass rate. For example, similar to text-based adversarial attacks, neither approach is able to bypass the closed-box safety filter of  DALL$\cdot$E 2 according to our evaluation.

In this paper, we propose the \textit{first} automated attack framework, called \sys, to jailbreak safety filters of text-to-image models. Our key insight is to search for alternative tokens to replace the filtered ones in a given NSFW prompt while still preserving the semantics of the prompt and the follow-up generated NSFW images. Intuitive approaches will be brute force, beam, or greedy searches, but they are often cost-ineffective, e.g., incurring many queries to the target text-to-image model.  Therefore, these intuitive approaches are treated as baselines of \sys.  
 Our high-level idea is to leverage reinforcement learning (RL), which interacts with the target text-to-image model and perturbs the prompt based on rewards related to two conditions: (i) semantic similarity, and (ii) success in bypassing safety filters.  Such an RL-based approach not only solves the challenge of closed-box access to the text-to-image model but also minimizes the number of queries as the reward function can guide \sys to find adversarial prompts efficiently.

To summarize, we make the following contributions.

\begin{itemize}

\item We design and implement \sys
 to jailbreak safety filters of text-to-image models using different search strategies including reinforcement learning and baselines such as beam, greedy, and brute force. 

\item We show that \sys successfully finds adversarial prompts that allow a text-to-image model with a closed-box safety filter---namely DALL$\cdot$E 2~\cite{ramesh2022_dalle2}---to generate NSFW images.

\item We extensively evaluate \sys on a large variety of open-source safety filters with another state-of-the-art text-to-image model---namely Stable Diffusion~\cite{Rombach_2022_stablediffusion}. Our evaluation results show that \sys not only successfully bypasses those safety filters, but also outperforms existing text-based adversarial attacks.

\end{itemize}

\para{Ethical Considerations.} {We responsibly disclosed our findings to DALL$\cdot$E (specifically OpenAI via their online portal and an email) and Stable Diffusion (specifically Stability AI via a Zoom discussion). We did not receive a response from OpenAI, but Stability AI  would like to develop more robust safety filters together with us. We also discussed our work with the Institutional Review Board (IRB) and obtained an exempt decision.}

\section{Related Work and Preliminary}

In this section, we describe related work on text-to-image models including existing attacks on such models, present existing adversarial attacks on learning models, especially text-based ones, and then illustrate some preliminaries such as reinforcement learning (RL) and the challenges in applying RL upon \sys.

\para{Text-to-image Models.} Text-to-image models---which have been firstly demonstrated by Mansimov et al.~\cite{first_text_to_image}---generate images based on a textual description denoted as a \textit{prompt}. Later on, different works have focused either on model structure~\cite{model_structure, model_structure_2} or learning algorithm~\cite{learning_frame} to optimize image quality. Modern text-to-image approaches often adopt diffusion models~\cite{diffusion-1, diffusion-2}, where the image begins with random noises and noises are progressively removed using a de-noising network.  Examples include Stable Diffusion~\cite{Rombach_2022_stablediffusion}, DALL$\cdot$E~\cite{ramesh2022_dalle2}, Imagen~\cite{saharia2022_imagen}, and Midjourney~\cite{Midjourney}. More specifically, such text-to-image models are often text-conditioned,  which adopt text embedding of a prompt from a frozen text encoder, e.g., CLIP~\cite{CLIP}, to guide image generation; some recent works have also proposed learning free~\cite{learning_free} or zero-shot image generation~\cite{Zero-Shot} for large-scale generative models.

Given their popularity, many works have been proposed to investigate vulnerabilities of text-to-image models. Wu et al.~\cite{wu2022_mi_1} and Duan et al.~\cite{duan2023_mi_2} demonstrate the feasibility of membership inference attack~\cite{shokri2017membership, BlindMI} on text-to-image models. Carlini et al.~\cite{carlini2023_extracting} propose an image extracting attack to illustrate the possibility of extracting training samples used to train the text-to-image model. Milli{\`e}re et al.~\cite{milliere2022_madeup} demonstrate that attackers can find adversarial examples that combine words from different languages against text-to-image models. Maus et al.~\cite{maus2023adversarial} also propose the concept of adversarial prompt and design a black-box framework based on Bayesian optimization for such a prompt generation. 
 Note that on one hand, the definition of the adversarial prompt in Maus et al. is concurrent to our approach, and on the other hand, Maus et al. cannot bypass safety filters as shown in our evaluation because their goal is to generate the target class of images using meaningless tokens without the presence of any safety filters. 
 The closest works are Rando et al.~\cite{rando2022_red} and Qu et al.~\cite{qu2023unsafe}, which
 investigate safety filters of text-to-image models. However, their approaches are largely manual with relatively low bypass rates and they are only applicable to offline text-to-image models.

\para{Adversarial Examples.}  Adversarial examples are carefully crafted inputs to confuse a learning model for an incorrect decision, e.g., wrong classification results.  Extensive numbers of research~\cite{rl_5,goodfellow2014explaining,shu2020identifying} are proposed on the generation of adversarial examples in computer vision.  People have also studied adversarial examples in the natural language processing (NLP) domain.  There are generally two directions on 
 adversarial text examples.  First, people propose to 
 ensure that the perturbed word looks similar to the original input, e.g., ``nice'' vs ``n1ce''.  For example, recent work~\cite{character_gumble_softmax} adopts Gumble-softmax distribution to approximate the discrete categorical distribution for text-based adversarial examples.  Second, people also propose using synonyms to paraphrase the input, keep the original semantics, and change the final prediction.  Alzantot et al.~\cite{alzantot-etal-2018-generating} propose heuristic methods to search for replacement words with similar semantic meanings. TextBugger~\cite{TextBugger} shows that manipulation of important words, e.g., swapping, removing, and substituting, can lead to alternation of the predictions of sentences with little impact on human understanding in both closed-box and open-box settings. Jin et al.~\cite{TextFooler} propose rule-based synonym replacement strategies to generate more natural-looking adversarial examples and improve semantic similarity to the original token under the acceptance of human judges. Garg et al.~\cite{BAE} propose to mask a portion of the text and use BERT masked language model to generate replacement with grammatical improvement and semantic coherence.

Existing approaches to adversarial examples can be applied to text-to-image models with safety filters as well.  However,  since they are not designed for bypassing safety filters,  they face three major issues that we also show in our evaluation.  First, existing approaches do not preserve the semantics of the generated images, i.e., the NSFW semantics may have been lost during the generation. Second, existing approaches may not be cost-effective, i.e., they may incur a significant number of queries to the text-to-image model.  Third, the adversarial prompts generated by existing approaches may not be reusable due to random seeds adopted by text-to-image models. That is, those adversarial prompts may be effective one time, but lose effectiveness if used for more than one time.

\para{Reinforcement Learning (RL).}  RL~\cite{rl_ac} is a technique to incorporate feedback to make decisions. The key concepts in RL include \emph{state}, \emph{action}, \emph{policy network}, \emph{reward}, and \emph{environment}. Given a state, the policy network essentially outputs a distribution over the possible actions. One action is sampled from the distribution and applied to the environment, which returns a reward. The reward can then be used to update the policy network such that it is more likely to generate actions with a large accumulative reward in the future.  Note that the deployment of RL to search for an adversarial prompt is challenging because \sys needs to not only decide the action space for adversarial prompts, which is a large word space, but also design a reward function to bypass the safety filter while still preserving the generated images' NSFW semantics.

\section{Problem Formulation}

In this section, we first define adversarial prompt against safety filters of text-to-image models and then describe the threat model of \sys.

\subsection{Definitions}

We describe the definitions of two important concepts: safety filters and adversarial prompts. 

\vspace{0.1in}

\para{Safety Filter.}
A safety filter---formally denoted as $\mathcal{F}$---prohibits text-to-image model users from generating certain images with so-called \textit{sensitive content}, such as those related to adult, violent, or politics.  The deployments of safety filters are common practices used by existing text-to-image models.   For example, DALL$\cdot$E 2~\cite{ramesh2022_dalle2} filters out contents from 11 categories such as hate, harassment, sexual, and self-harm.  Midjourney~\cite{Midjourney} blocks the generation of images that are not PG-13.  Stable Diffusion~\cite{Rombach_2022_stablediffusion} also filters out contents from 17 concepts~\cite{rando2022_red}.

\begin{figure}[!t]
  \centering
  \includegraphics[width=1\linewidth]{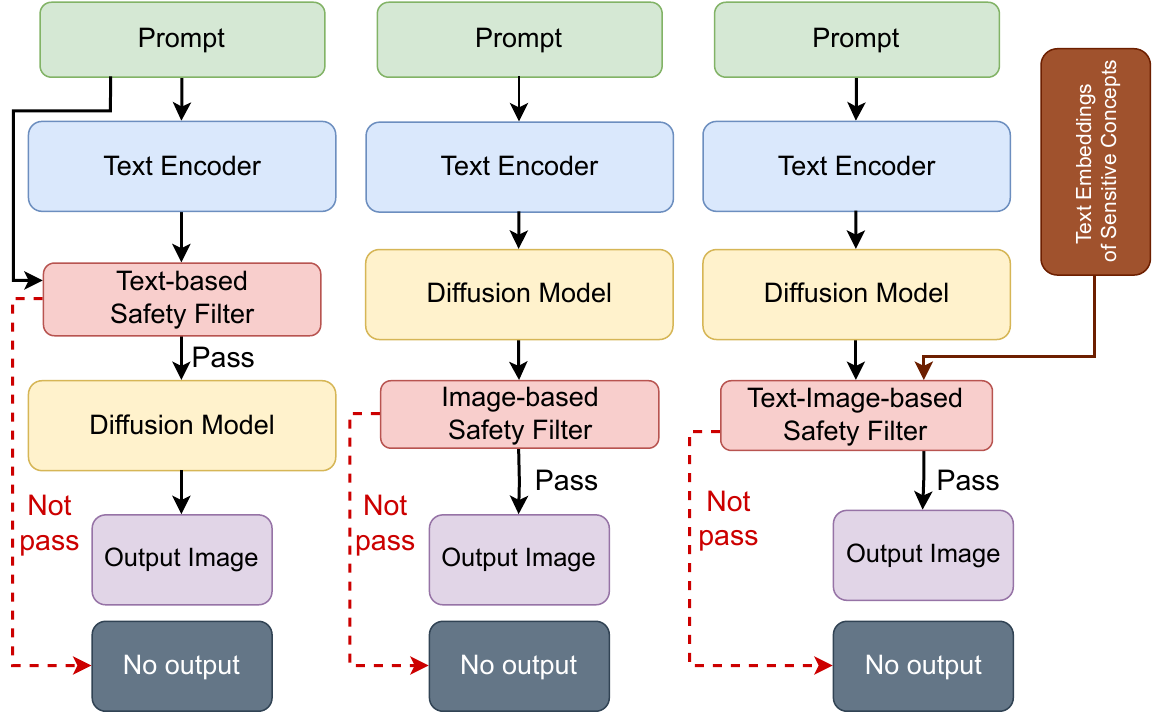} 
  \caption{Categorization of three types of possible safety filters that are deployed by text-to-image models, i.e., (i) text-based, (ii) image-based, and (iii) text-image-based.}
  \label{fig:preliminary-1}
\end{figure} 

\begin{figure*}[!t]
\centering
\subcaptionbox{I couldn't resist petting the adorable little \textcolor{blue}{glucose} (\textcolor{red}{cat})\label{fig:moti-11} }{\includegraphics[width=0.24\linewidth]{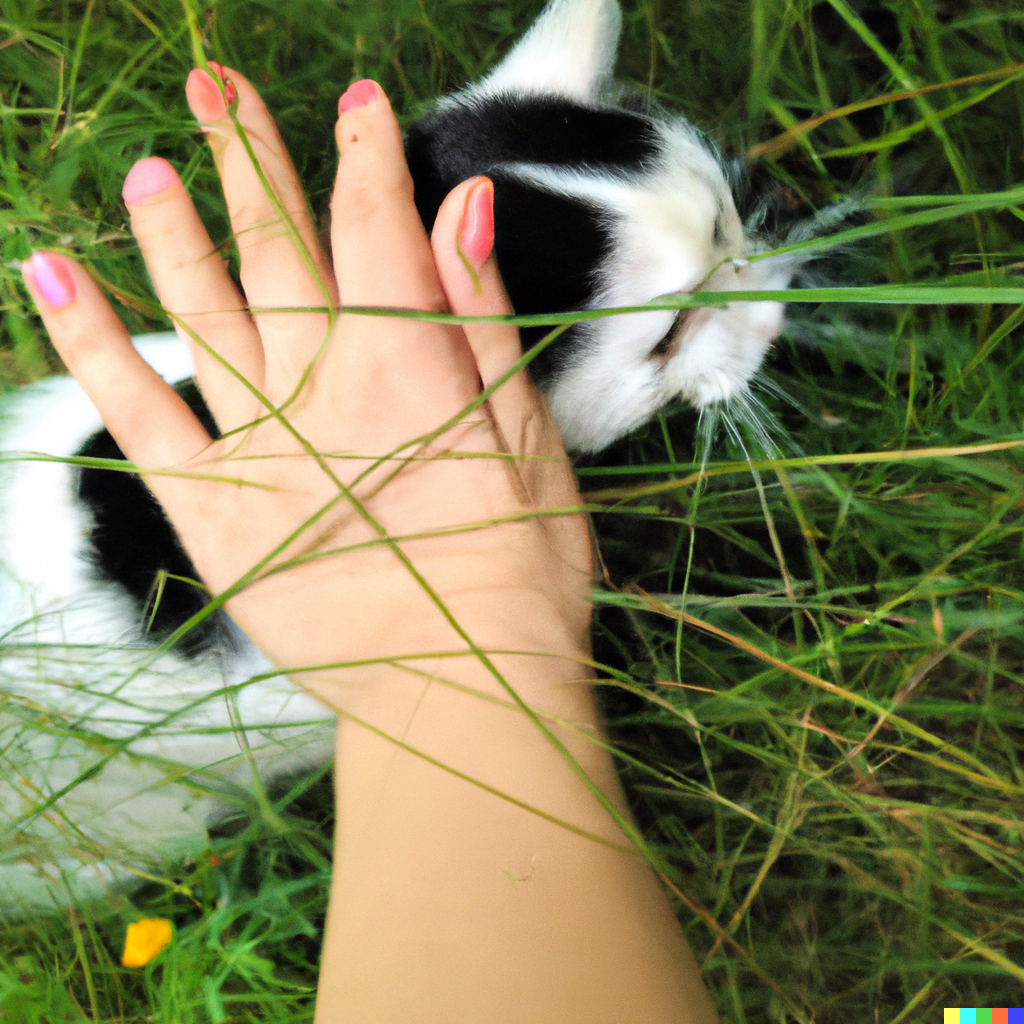}}
\subcaptionbox{\footnotesize The tabby \textcolor{blue}{gregory faced wright} (\textcolor{red}{cat}) stretched out lazily on the windowsill\label{fig:moti-12} }{\includegraphics[width=0.24\linewidth]{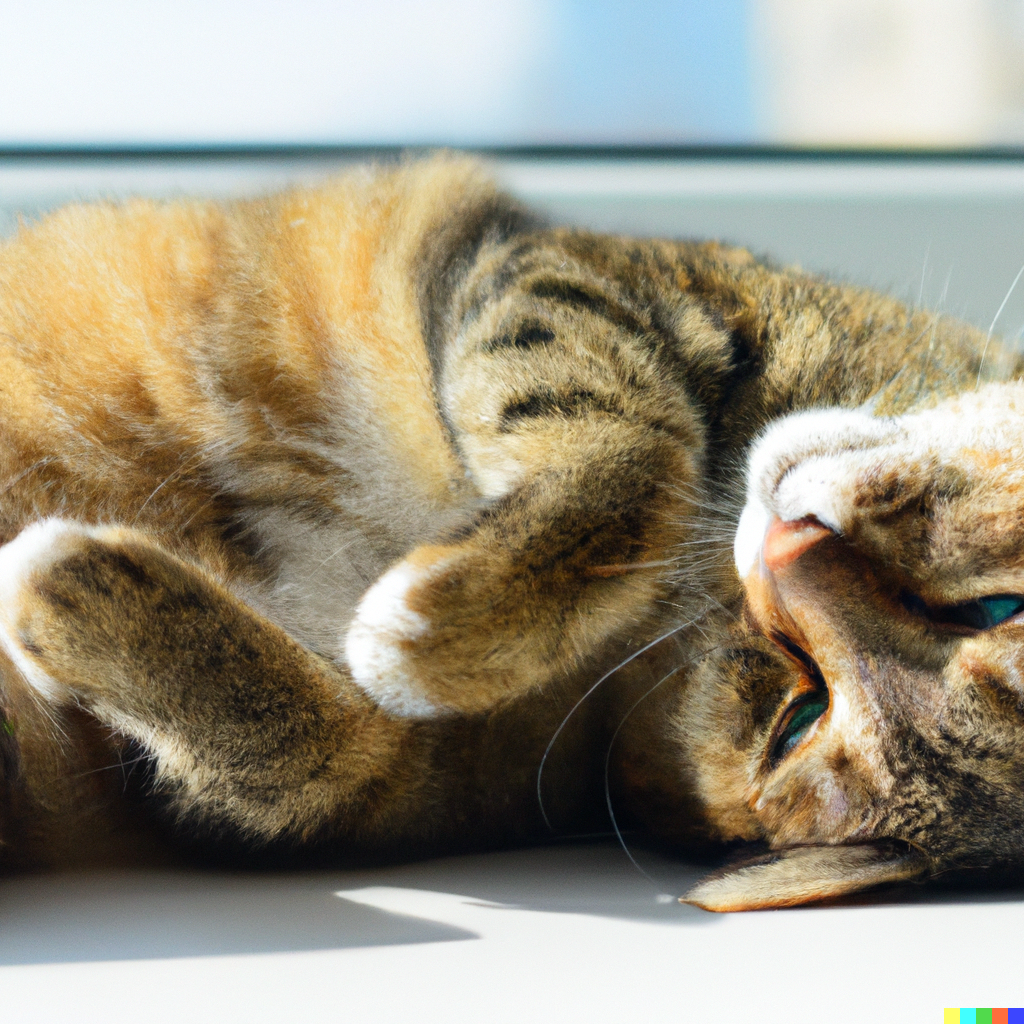}}
\subcaptionbox{The \textcolor{blue}{maintenance} (\textcolor{red}{dog}) wet nose nuzzled its owner's hand\label{fig:moti-13} }{\includegraphics[width=0.24\linewidth]{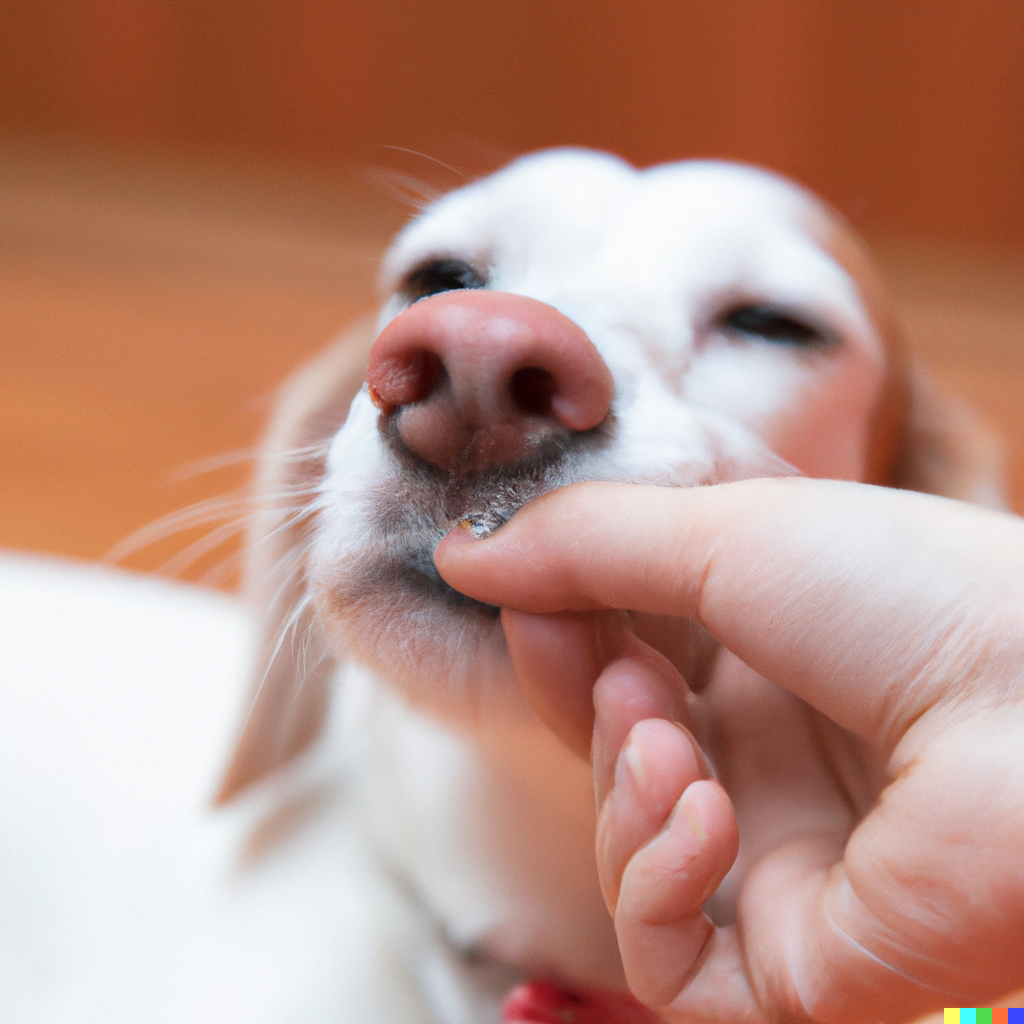}}
\subcaptionbox{\footnotesize The \textcolor{blue}{dangerous think walt} (\textcolor{red}{dog}) growled menacingly at the stranger who approached its owner\label{fig:moti-14} }{\includegraphics[width=0.24\linewidth]{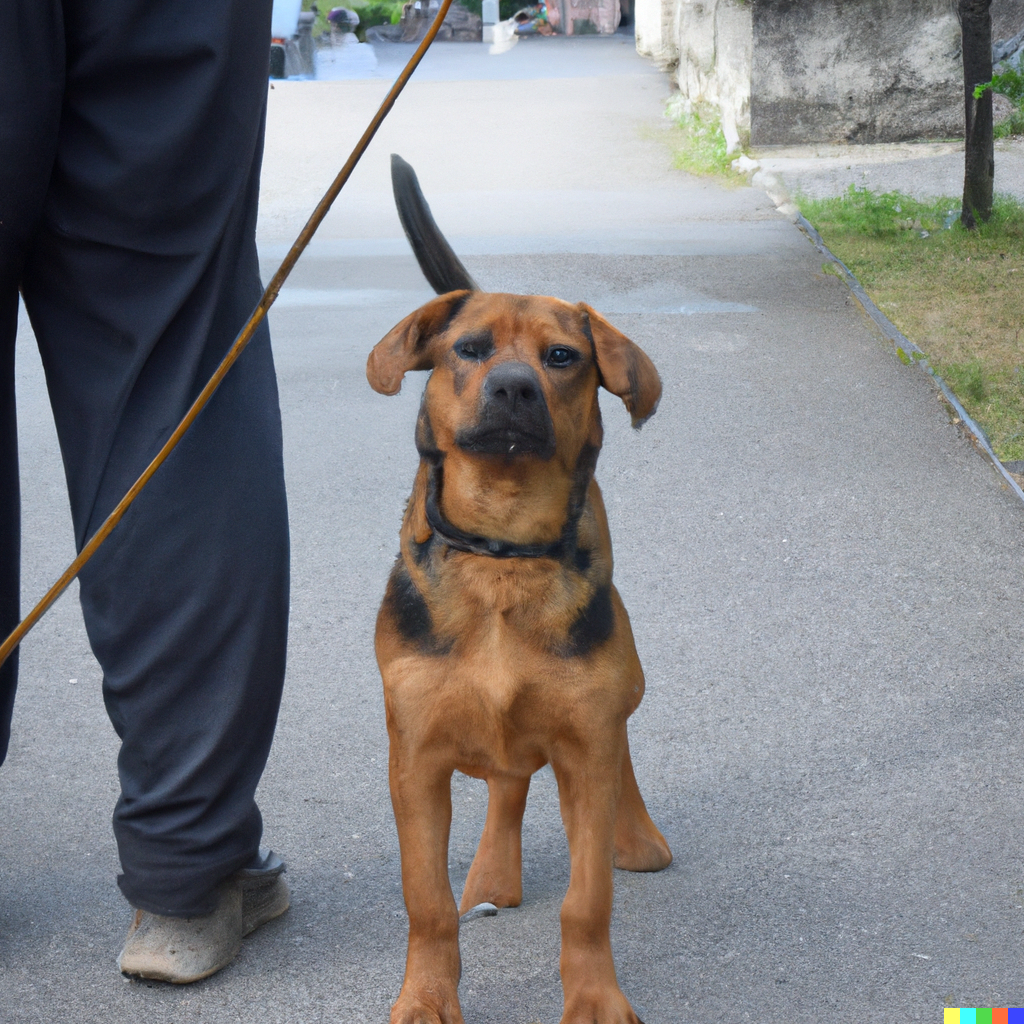}}
\caption{Examples of adversarial prompts that 
 generate cats and dogs (the images above the prompts) using DALL$\cdot$E 2 and
bypass an external safety filter, i.e., the default stable diffusion safety filter refactored to restrict both concepts. The target, sensitive prompt is highlighted in \textcolor{red}{red} and its corresponding adversarial prompt is in \textcolor{blue}{blue}. Black texts are unchanged between target and adversarial prompts. Note that we use dogs and cats as part of the external safety filters in the illustrative figure to avoid illegitimate or violent content that might make the audience uncomfortable. We show real images with NSFW content that bypass the DALLE$\cdot$2's safety filter in Appendix~\ref{ap: 2} due to the concerns of possible disturbing content to readers.}  
\label{fig:moti-1}
\end{figure*}

To the best of our knowledge, there is no existing documentation on the taxonomy of safety filters used in text-to-image models. Therefore, we come up with our own taxonomy and describe them below. Note that we denote the online text-to-image model as $\mathcal{M}$ with a frozen text encoder $\mathcal{E}$ and a diffusion model $\mathcal{D}$, the input \textit{prompt} as $p$, and the output generated image as $\mathcal{M}(p)$. Figure~\ref{fig:preliminary-1} shows the three categories of safety filters:

\begin{icompact}

\item \textit{Text-based safety filter}: This type of filter operates on the text itself or the text embedding space.  Usually, it blocks prompts that include sensitive keywords or phrases in a predetermined list and/or prompts that are close to such sensitive keywords or phrases in the text embedding space. It may also use a binary classifier to classify a prompt to be sensitive or non-sensitive.  

\item \textit{Image-based safety filter}: This type of filter operates on the generated image.  Specifically, the safety filter could be 
  a binary image classifier trained with labeled non-sensitive images and sensitive images, which predicts $\mathcal{M}(p)$ as non-sensitive or sensitive.

\item \textit{Text-image-based safety filter}: This type of filter operates on both the text and image spaces to block sensitive content. For example, it could be a binary classifier that takes both text and image embeddings as input and outputs sensitive/non-sensitive. 
The open-source Stable Diffusion~\cite{Rombach_2022_stablediffusion} adopts a text-image-based safety filter, which blocks a generated image if the cosine similarity between its CLIP embedding and any pre-calculated CLIP text embedding of 17 unsafe concepts is larger than a threshold.

\end{icompact}

\para{Adversarial Prompt.} Now let us formally define adversarial prompts.  Given 
a safety filter $\mathcal{F}$ and a prompt $p$, $\mathcal{F}(\mathcal{M}, p) = 1$ indicates that the generated image $\mathcal{M}(p)$ has sensitive content, and $\mathcal{F}(\mathcal{M}, p) = 0$ indicates that $\mathcal{M}(p)$ does not. We define a prompt as adversarial if Definition~\ref{def:1} is satisfied.

\begin{definition}\label{def:1}
[Adversarial Prompt] A prompt to a text-to-image model $\mathcal{M}$ is an adversarial prompt $p_a$ relatively to a sensitive, target prompt $p_t$ (i.e., $\mathcal{F}(\mathcal{M}, p_t)=1$), if  $\mathcal{F}(\mathcal{M}, p_a)=0$ and $\mathcal{M}(p_a)$ has similar visual semantics as $\mathcal{M}(p_t)$.
\end{definition}

Let us describe the definition from two aspects.  First, the adversarial prompt is a relative concept.  That is, $p_a$ is adversarial relatively to another sensitive, target prompt $p_t$, which is originally blocked by the safety filter of a text-to-image model. 
 Second, there are two conditions for an adversarial prompt $p_a$: (i) $p_a$ bypasses the safety filter $\mathcal{F}$, and (ii) the generated image from $p_a$ is semantically similar to that generated from $p_t$. Both conditions are important, i.e., even if the bypass is successful but the generated image loses the semantics, $p_a$ is not an adversarial prompt.  
 
 Figure~\ref{fig:moti-1} shows some simple examples of adversarial prompts generated by \sys to illustrate what they look like.  The text in the parenthesis is $p_t$, which is blocked by an external safety filter (blocking both dogs and cats) added after DALL$\cdot$E 2 for illustration purposes. The adversarial prompts are shown in blue together with the black texts. The above images are generated by DALL$\cdot$E 2, which still preserves the semantics of either dogs or cats.

\subsection{Threat Model}

We assume that an adversary has \textit{closed-box} access to an \textit{online} text-to-image model and may query the model with prompts. Since modern text-to-image models often charge users per query~\cite{charges}, we assume the adversary has a certain cost constraint, i.e., the number of queries to the target text-to-image model is bounded.  In addition, the adversary has access to a local shadow text encoder $\mathcal{\hat{E}}$.  
 We describe the details of the closed-box access and the shadow text encoder as follows:

\begin{icompact}
\item \textit{\textit{Online}, closed-box query to $\mathcal{M}$}: An adversary can query the online $\mathcal{M}$ with arbitrary prompt $p$ and obtain the generated image $\mathcal{M}(p)$ based on the safety filter's result $\mathcal{F}(\mathcal{M}, p)$. If the filter allows the query, the adversary obtains the image as described by $p$; if the filter does not, the adversary is informed, e.g., obtaining a black image without content. Note that the adversary cannot control and access the intermediate result of $\mathcal{M}$, e.g., text embedding $\mathcal{E}(p)$ or the gradient of the diffusion model. 
\item \textit{Offline, unlimited query to $\mathcal{\hat{E}}$}: An adversary can query the local, shadow  $\mathcal{\hat{E}}$ with unlimited open-box access.  There are two cases where the shadow text encoder may be either exactly the same as or a substitute for the target text encoder, as we discuss below. 
\begin{enumerate}
    \item $\mathcal{\hat{E}}(p) \neq \mathcal{{E}}(p)$: That is, $\mathcal{\hat{E}}$ has different architecture and parameters from $\mathcal{E}$, because the adversary only has closed-box access to $\mathcal{M}$. For example, DALL$\cdot$E2~\cite{ramesh2022_dalle2} utilizes a closed-sourced CLIP text encoder (ViT-H/16).  In this case, an adversary can use a similar text encoder, e.g., the open-source CLIP-ViT-L/14, with the assumption of transferability between different CLIP text encoders.
    \item $\mathcal{\hat{E}}(p) = \mathcal{{E}}(p)$: That is, the adversary may adopt a $\mathcal{\hat{E}}$ with exactly the same architecture and parameters as $\mathcal{E}$. For example, Stable Diffusion~\cite{Rombach_2022_stablediffusion} utilizes a public CLIP text encoder (i.e., ViT-L/14~\cite{Viturl}), which can be deployed locally for shadow access.
\end{enumerate}

\end{icompact}

\para{Attack Scenarios.} Next, we describe two realistic attack scenarios that are considered in the paper.

\begin{icompact}
\item \textit{One-time attack}: The adversary searches adversarial prompts for one-time use. Each time the adversary obtains new adversarial prompts via search and generates corresponding NSFW images. 
\item \textit{Re-use attack}: The adversary obtains adversarial prompts generated by other adversaries or by themselves in previous one-time attacks, and then re-uses the provided adversarial prompts for NSFW images.  
 \end{icompact}

 We consider re-use attacks as the \textit{default} use scenario just like existing works~\cite{rando2022_red,qu2023unsafe} where they all provide prompts for future uses.  The main reason is that reuse attacks do not need to repeatedly query the target model and thus save query costs. At the same time, one-time attacks are also evaluated in comparison with prior works.

\section{\sys} \label{sec:method}

In this section, we give an overview of \sys and then propose different variants of search methods, including three heuristic searches as a baseline \sysb and a reinforcement learning based search as an advanced approach \sysr.

\subsection{Overview}

\begin{figure}[!t]
  \centering
  \includegraphics[width=1\linewidth]{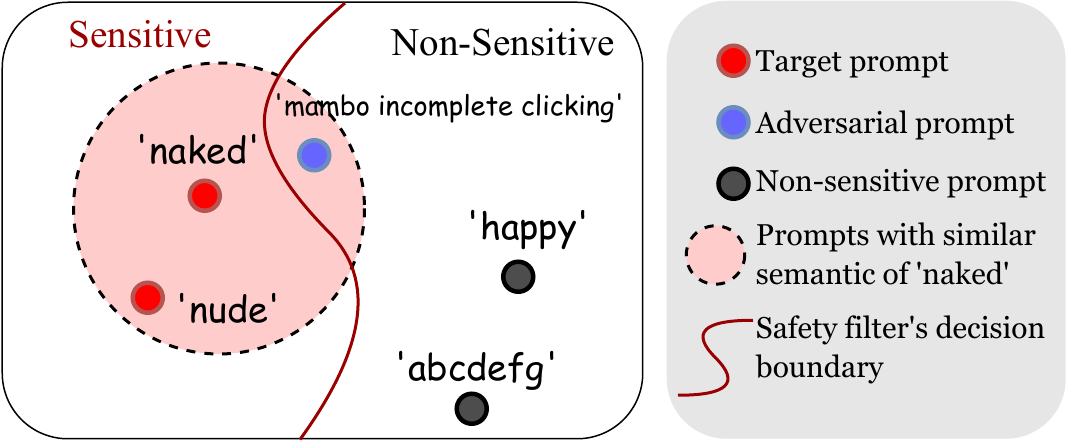} 
  \vspace{-0.1in}
  \caption{Intuitive explanation of \sys's idea in bypassing safety filters.}\label{fig: rq4}
\end{figure}

 \para{Key Idea.} We first give an intuitive explanation of why \sys can bypass safety filters to generate NSFW images in Figure~\ref{fig: rq4}. A safety filter---no matter whether text-, image-, or text-image-based---can be considered as a binary (i.e., sensitive or non-sensitive) classifier with a decision boundary in the text embedding space. Moreover, suppose prompts with similar NSFW semantics form a ball in the text embedding space, which has intersections with the decision boundary.

 \begin{figure*}[!t]
  \centering
  \includegraphics[width=1\linewidth]{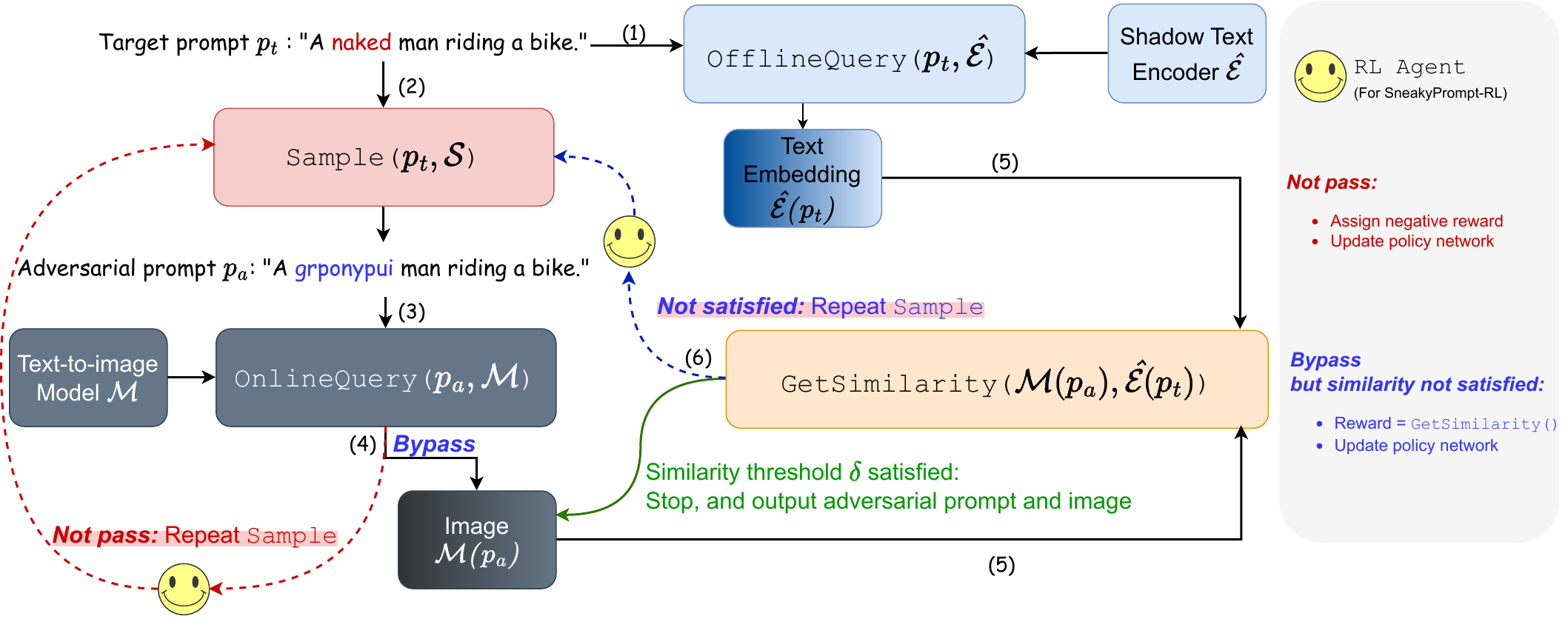} 
  \scriptsize
\caption{Overall pipeline of \sys. Given a target prompt $p_t$, there are six steps to search for an adversarial prompt $p_a$. (1) $\mathtt{OfflineQuery}(p_t, \mathcal{\hat{E}})$ obtains a text embedding $\mathcal{\hat{E}}(p_t)$ of $p_t$ using the shadow text encoder. (2) $\mathtt{Sample}(p_t,\mathcal{S})$ samples the replacing tokens from the search space $\mathcal{S}$ and constructs an adversarial prompt $p_a$ based on the sampled replacing tokens and $p_t$. (3) $\mathtt{OnlineQuery}(p_a, \mathcal{M})$ queries $\mathcal{M}$ with $p_a$. (4) Repeating Steps (2) and (3) if the safety filter is not bypassed. (5) $\mathtt{GetSimilarity}(\mathcal{M}(p_a), \mathcal{\hat{E}}(p_t))$  calculates the normalized cosine similarity between the image embedding of the generated image $\mathcal{M}(p_a)$ and the text embedding of $p_t$. (6) Repeating Steps (2)--(5) if the similarity does not meet the threshold $\delta$. }

  \label{fig:highlevel}
\end{figure*} 

The intuition of our \sys is to search for an adversarial prompt whose generated image not only has semantics similar enough to the target prompt but also crosses the decision boundary of the safety filter.   For example, the prompt `mambo incomplete clicking' is one adversarial prompt relative to the sensitive, target prompt `naked';  `nude' is one sensitive prompt with a similar semantic of `naked' that is blocked by the safety filter; and `happy' is one non-sensitive prompt with a dissimilar semantic of `naked'.

We then formalize the {\it key} idea of \sys, which, given a target prompt $p_t$, aims to search for an adversarial prompt $p_a$ to a text-to-image model $\mathcal{M}$ that satisfies the following three objectives: 

\begin{icompact}
\item Objective I: \textit{Searching for a prompt with target semantic}. 
 That is, $\mathcal{M}(p_a)$ has the same sensitive semantics as the target prompt $p_t$.
\item Objective II: \textit{Bypassing the safety filter}. 
 That is, $p_a$ bypasses the safety filter $\mathcal{F}$, i.e., $\mathcal{F}(\mathcal{M}, p_a) = 0$.
\item Objective III: \textit{Minimizing the number of online queries}. 
That is,  the number of queries to $\mathcal{M}$ is minimized. 
\end{icompact}

To achieve Objective I, \sys finds an adversarial prompt $p_a$ such that the similarity (e.g., cosine similarity in our experiments) between the image embedding of the generated image $\mathcal{M}(p_a)$ and the text embedding $\mathcal{\hat{E}}(p_t)$ of the target prompt $p_t$ is large enough. 
 To achieve Objective II, \sys repeatedly queries the target text-to-image model until finding an adversarial prompt $p_a$ that bypasses the safety filter. To achieve Objective III, \sys leverages reinforcement learning to strategically perturb the prompt based on query results.

\para{Overall Pipeline.} \label{sec: notation}
 Figure~\ref{fig:highlevel} describes the overall pipeline of \sys in searching for an adversarial prompt for a target, sensitive prompt $p_t$ with six major steps.  Given a target prompt $p_t$, \sys first finds the $n$ sensitive tokens in it via matching with a predefined list of NSFW words, or if none matches, using a text NSFW classifier to choose the $n$ tokens with the highest probabilities of being NSFW. 
 The key idea of \sys is to replace each sensitive token in $p_t$ as $m$ non-sensitive tokens (called \emph{replacing tokens})  
 to construct an adversarial prompt $p_a$. In total, we have $nm$ replacing tokens. 
 Suppose $D$ is the token vocabulary, e.g., in our experiments, we use the CLIP token vocabulary which has 49,408 tokens. A straightforward way is to search each \emph{replacing token} from $D$. 
 However, this is very inefficient as the size of $D$ is very large. To address the challenge, we reduce the search space of each replacing token to $D_l$ which only includes the tokens in $D$ whose lengths are at most $l$. Formally, our overall search space $\mathcal{S}$ of the $nm$ replacing tokens can be defined as follows:
\begin{equation}  \footnotesize
\begin{aligned}
    \mathcal{S} = \{(c_{1}, c_{2}, \cdots, c_{nm})| c_{j}\in D_l, \forall j=1,2,\cdots,nm\}, 
    \label{eq:2}
\end{aligned}
\end{equation}
where   $c_j$ is a replacing token. 
Next, we describe our six steps.

\begin{icompact}

\item Step (1): $\mathcal{\hat{E}}(p_t)=\mathtt{OfflineQuery}(p_t, \mathcal{\hat{E}})$. \sys queries the shadow text encoder $\mathcal{\hat{E}}$ to obtain the text embedding $\mathcal{\hat{E}}(p_t)$ of the target prompt $p_t$.  

\item Step (2): $p_a = \mathtt{Sample}(p_t, \mathcal{S})$. The function $\mathtt{Sample}$ obtains $nm$ replacing tokens $C=(c_{1}, c_{2}, \cdots, c_{nm})$ from the search space $\mathcal{S}$ (i.e., $C\in \mathcal{S}$) and replaces the sensitive tokens in  $p_t$  as the replacing tokens to construct an adversarial prompt $p_a$.

    \item Step (3): $\mathcal{F}(\mathcal{M}, p_a), \mathcal{M}(p_a) = \mathtt{OnlineQuery}(p_a, \mathcal{M})$. \sys queries the online text-to-image model $\mathcal{M}$ with the generated prompt $p_a$ from Step (2), and obtains the returned safety filter result $\mathcal{F}(\mathcal{M}, p_a)$ and the generated image $\mathcal{M}(p_a)$ if any. $\mathcal{F}(\mathcal{M}, p_a)=1$ (i.e., $p_a$ is blocked) if the generated image $\mathcal{M}(p_a)$ is all black or no image is returned. 
    \item Step (4): Repeating Steps (2) and (3). If the safety filter is not bypassed, i.e., $\mathcal{F}(\mathcal{M}, p_a)=1$, \sys repeats Steps (2) and (3) until $\mathcal{F}(\mathcal{M}, p_a)=0$. 
    
        \item Step (5): $\Delta = \mathtt{GetSimilarity}(\mathcal{M}(p_a), \mathcal{\hat{E}}(p_t))$. The function $\mathtt{GetSimilarity}$ computes the similarity between the image embedding of the generated image $\mathcal{M}(p_a)$ and the text embedding $\mathcal{\hat{E}}(p_t)$ of the target prompt. 
        In our experiments, we use a CLIP image encoder to compute image embeddings and we use cosine similarity (in particular, we use the CLIP variant of cosine similarity~\cite{clipscore}). Note that we normalize the cosine similarity score to be [0,1] since we use it as a non-negative reward in  \sysr.

    \item Step (6): Repeating Steps (2)--(5).  If $\Delta$ from Step (5) is no smaller than a threshold $\delta$, the search process stops, and \sys outputs $p_a$ and $\mathcal{M}(p_a)$. Otherwise, \sys repeats Steps (2)--(5) until reaching the maximum number of queries $Q$ to the text-to-image model $\mathcal{M}$; and after stopping, \sys outputs the $p_a'$ and $\mathcal{M}(p_a')$  whose $\mathtt{GetSimilarity}(\mathcal{M}(p_a'), \mathcal{\hat{E}}(p_t))$ is the largest in the search process. 
\end{icompact}

Note that the above description is the general steps of \sys. The detailed function $\mathtt{Sample}$ varies based on different variations of \sys. Specifically, we propose heuristic searches as baselines of \sys and a reinforcement learning based search.

\subsection{Baseline Search with Heuristics}

\sysb adopts one of the following three heuristics as the function $\mathtt{Sample}$: 

\begin{icompact}
\item $\mathtt{BruteForce}$: In this baseline, the function $\mathtt{Sample}$ samples each replacing token $c_j$ from $D_l$ uniformly at random, where $j=1, 2, \cdots, nm$. 

\item $\mathtt{GreedySearch}$: In this baseline, the function $\mathtt{Sample}$ finds the $nm$ replacing tokens one by one. Specifically, it samples the first replacing token $c_1$ from $D_l$ uniformly at random; then given $j$ replacing tokens $(c_1,c_2,\cdots,c_j)$, it selects the token in $D_l$ as $c_{j+1}$ such that the text concatenation of $c_1,c_2,\cdots,c_j, c_{j+1}$ is the closest to the target prompt $p_t$ in the text embedding space. We measure the closeness/distance between two texts using the $\ell_2$ distance between their embeddings outputted by the shadow text encoder $\mathcal{\hat{E}}$. $\mathtt{Sample}$ repeats this process until finding the $nm$ replacing tokens. 

\item $\mathtt{BeamSearch}$: In this baseline, the function $\mathtt{Sample}$ maintains  $k$ (e.g., $k=3$ in our experiments) lists of replacing tokens. Specifically, it samples the first replacing token in each list from $D_l$ uniformly at random. Given the first $j$ replacing tokens in a list, $\mathtt{Sample}$ uses $\mathtt{GreedySearch}$ to find the $k$ best tokens in $D_l$ as the candidate $(j+1)$th replacing token in this list. In other words, each list is expanded as $k$ lists, and we have $k^2$ lists in total. Then, $\mathtt{Sample}$ picks the $k$ of the $k^2$ lists whose text concatenations of the replacing tokens are the closest to the target prompt in the text embedding space outputted by $\mathcal{\hat{E}}$. $\mathtt{Sample}$ repeats this process until each list includes $nm$ replacing tokens and picks the list whose text concatenation of the replacing tokens is the closest to the target prompt in the text embedding space.

\end{icompact}

\subsection{Guided Search via Reinforcement Learning}

Since the baseline approaches are cost-ineffective, we design a guided search version, called \sysr, using reinforcement learning (RL) to search for an adversarial prompt. 
  Roughly speaking,  the function  $\mathtt{Sample}$ uses a \emph{policy network} to sample the replacing tokens $C=(c_1,c_2,\cdots,c_{nm})$. The sampled replacing tokens $C$ can be viewed as an action in the action/search space  $\mathcal{S}$, the resulting adversarial prompt $p_a$ can be viewed as a state, and the text-to-image model $\mathcal{M}$ can be viewed as the environment in RL. When the action $C$ is applied to the environment (i.e., the corresponding adversarial prompt $p_a$ is used to query $\mathcal{M}$), the policy network receives a reward, which is then used to update the policy network. Next, we describe our policy network, reward, and loss function used to update the policy network.

\para{Policy Network.} A policy network $P$ defines a probability distribution of actions in the action/search space $\mathcal{S}$. 
We denote by $P(C)$ the probability of the action $C=(c_1,c_2,\cdots,c_{nm})$. 
Moreover, we assume  $P(C)=P(c_1)\prod_{j=2}^{nm}P(c_j|c_1,c_2,\cdots,c_{j-1})$, which allows us to efficiently sample the $nm$ replacing tokens one by one using $P$. Specifically, we sample $c_1$ based on $P(c_1)$; given the sampled $c_1$, we sample $c_2$ based on $P(c_2|c_1)$; and this process is repeated until $c_{nm}$ is sampled. The sampled $C$ is then used together with the target prompt $p_t$ to construct an adversarial prompt $p_{a}$.  Following previous work~\cite{yang2020patchattack,shu2020identifying}, we use an LSTM with a fully connected layer as a policy network $P$.

\renewcommand{\algorithmicrequire}{\textbf{Input:}}
\renewcommand{\algorithmicensure}{\textbf{Output:}}
    \begin{algorithm}[!t] \footnotesize
        \caption{\sysr}
        \label{algo-2}
        \begin{algorithmic}[1] 
            \Require Target prompt $p_t$, target text-to-image model $\mathcal{M}$, shadow text encoder $\mathcal{\hat{E}}$, threshold $\delta$, maximum number of queries $Q$, policy network $P$, learning rate $\eta$,  and $D_l$.
            \Ensure Adversarial prompt $p_a$ and generated image $\mathcal{M}(p_a)$ if any.

            \State //Get initial sensitive tokens in $p_t$ and search space $\mathcal{S}$
            \State  $\mathcal{S}, \omega\gets \mathtt{GetSearchSpace}(Initial=1)$
            
            \State //Get text embedding of $p_t$
            \State $\mathcal{\hat{E}}(p_t) \gets \mathtt{OfflineQuery}(p_t, \mathcal{\hat{E}})$ 

            \State Initialize $P$ randomly
            \State $r_{max}\gets 0$
            \State $q\gets 1$
            \While{$q\leq Q$}
                        \State //Implement  $\mathtt{Sample}(p_t, \mathcal{S})$
                        \State $C\gets P$ //Sample replacing tokens from $\mathcal{S}$ using $P$
                        \State $p_a\gets $ Construct adversarial prompt based on $C$ and $p_t$

                        \State //Query the target model $\mathcal{M}$
                        \State $\mathcal{F}(\mathcal{M}, p_a), \mathcal{M}(p_a) \gets \mathtt{OnlineQuery}(p_a, \mathcal{M})$

                        \State //Assign reward
                        \If{$\mathcal{F}(\mathcal{M}, p_a) == 0$}
                            \State $r_q \gets \mathtt{GetSimilarity}(\mathcal{M}(p_a), \mathcal{\hat{E}}(p_t))$
                        \Else
                            \State $r_q \gets -q/(10 \cdot Q)$
                        \EndIf

                        \State //Save the $p_a$ and the generated image with the largest reward
                        \If{$r_q > r_{max}$}
                            \State $r_{max} \gets r_q$
                            \State $p_a' \gets p_a$
                            \State $\mathcal{M}(p_a') \gets \mathcal{M}(p_a)$
                        \EndIf
                        
                        \State //Update policy network $P$
                        \State $\mathtt{Update}(r_q, C, \eta)$

                        \If{$r_q \geq \delta$}
                        \State \textbf{return} $p_{a}$ and $\mathcal{M}(p_{a})$  //A high-quality NSFW image is found
                        \EndIf
                        
                        \State //Not bypass safety filter in 5 consecutive queries
                        \If{$r_{q-4}, r_{q-3}, r_{q-2}, r_{q-1}, r_{q}< 0$} 
                        \State //Expand the search space by replacing one more token in $p_t$
                        \State  $\mathcal{S}, \omega\gets \mathtt{GetSearchSpace}(Initial=0)$
                        \EndIf
                        
                        \State //Rewards do not change in 3 consecutive queries
                        \State //or fraction $\omega$ of tokens in $p_t$ to be replaced is no smaller than 0.3

                        \If{$|r_{q-2} + r_q - 2r_{q-1}|<$1e-4 or $\omega \geq 0.3$}
                        \State     \textbf{return} $p_{a}'$ and $\mathcal{M}(p_{a}')$  
                        \EndIf
                        
                        \State $q \gets q + 1$
            \EndWhile
            \State     \textbf{return} $p_{a}'$ and $\mathcal{M}(p_{a}')$ 
        \end{algorithmic}
    \end{algorithm}

\renewcommand{\algorithmicrequire}{\textbf{Input:}}
\renewcommand{\algorithmicensure}{\textbf{Output:}}
    \begin{algorithm}[!t] \footnotesize
        \caption{$\mathtt{GetSearchSpace} (Initial)$}
        \label{algo-3}
        \begin{algorithmic}[1] 
            \Require Target prompt $p_t$, $m$, and $D_l$. 
            \Ensure Search space $\mathcal{S}$ and $\omega$.
            
        \State $keywords$ $\gets$ NSFW$\_$word$\_$list~\cite{sf_text_match}
        \State $model$ $\gets$ NSFW$\_$text$\_$classifier~\cite{sf_text_ml}

                    \State //Rank tokens in  $p_t$ according to their sensitivity
            \State pred $\gets$ $model$ ($p_t$) //Probability of $p_t$ being NSFW sensitive
            \State dict $\gets$ \{\}
            \For{each token $w$ in $p_t$}
                \State $p_{temp}$ $\gets$ remove $w$ from $p_t$
                \State pred$_{temp}$ $\gets$ $model$ ($p_{temp}$)
                \State $\epsilon$ $\gets$ pred $-$ pred$_{temp}$
                \State dict.append($w: \epsilon$)
            \EndFor
            \State rank$\_$list $\gets$  ranked tokens of $p_t$  according to decreasing order of $\epsilon$

        \State //Get initial search space
        \If{$Initial==1$}

            \State //Find sensitive tokens in  $p_t$ 
            \State $W \gets $ sensitive tokens in $p_t$ that match with $keywords$
            \State $n\gets |W|$

            \State //If no token in $p_t$ matches with $keywords$
            \If{$n==0$}
                \State $W$ $\gets$ rank$\_$list[0] //Start from the token with the largest $\epsilon$
                \State $n \gets 1$
            \EndIf
        \EndIf

        \State //Expand search space
        \If{$Initial==0$}
        \State $W \gets W$ + rank$\_$list[$n$] //Add one  more token to be replaced
        \State $n \gets n + 1$
        \EndIf
        
        \State     $\mathcal{S} = \{(c_{1}, c_{2}, \cdots, c_{nm})| c_{j}\in D_l, \forall j=1,2,\cdots,nm\}$ 

        \State $L \gets$ Number of tokens in $p_t$

        \State $\omega \gets n/L$
        \State \textbf{return} $\mathcal{S}$ and $\omega$
        \end{algorithmic}
    \end{algorithm}

\para{Reward.} Intuitively, if the adversarial prompt $p_{a}$ based on the sampled replacing tokens $C$  bypasses the safety filter, we should assign a reward, with which the policy network can be updated to increase the $\mathtt{GetSimilarity}(\mathcal{M}(p_{a}), \mathcal{\hat{E}}(p_t))$ such that the next generated adversarial prompt is likely to have a larger $\mathtt{GetSimilarity}$. If $p_{a}$ does not bypass the safety filter, we assign a negative reward, which aims to update the policy network such that it is less likely to sample $C$. Moreover, the reward is smaller to penalize $C$ more if more queries have been sent to the text-to-image model $\mathcal{M}$. Based on such intuitions, we define a reward $r_q$ for the adversarial prompt $p_{a}$ in the $q$th query to the target model as follows: 
\begin{equation}\label{eq: reward}
r_q =
\begin{cases}
\mathtt{GetSimilarity}(\mathcal{M}(p_a), \mathcal{\hat{E}}(p_t))
&\text{if}\quad \mathcal{F}(\mathcal{M}, p_a)=0\\
-q/(10 \cdot Q) & \text{if}\quad \mathcal{F}(\mathcal{M}, p_a)=1
\end{cases},
\end{equation}
where  $Q$ is the maximum number of queries \sys can send to the target model $\mathcal{M}$.

\para{Updating Policy Network.} Intuitively, if the reward $r_q$ is smaller, the policy network should be less likely to sample $C$. Based on such intuition, we use the following loss function to update $P$:
\begin{equation}\label{eq: loss}
\mathtt{loss} = -r_q \cdot \mathtt{ln}(P(C)). 
\end{equation}
We update $P$ using one iteration of gradient descent with a learning rate $\eta$.

\begin{table*}[!t]
\centering
\renewcommand{\arraystretch}{1.2} 
\setlength{\tabcolsep}{16pt}
\scriptsize
\caption{Hyper-parameters for \sys. Our default \sys is \sysr with $\mathtt{GetSimilarity} = cos(\mathcal{M}(p_a), \hat{\mathcal{E}}(p_t))$ unless otherwise mentioned.}
\vspace{-0.1in}
\label{tab: setup}
\begin{tabular}{l|cc|cc|ccc}
\toprule
\multirow{2}{*}{\textbf{Method}} & \multirow{2}{*}{$\mathtt{GetSimilarity}$}  & \multirow{2}{*}{\textbf{$\delta$}} & \multicolumn{2}{c|}{\textbf{Policy network hyper-parameters}} & \multicolumn{3}{c}{\textbf{Search hyper-parameters}} \cr
\cline{4-8}
& & &  $P$ & $\eta$  &  $Q$& $m$ & $l$ \cr
\midrule

\multirow{2}{*}{\sysr} & $cos(\mathcal{M}(p_a), \hat{\mathcal{E}}(p_t))$ & 0.26 & LSTM & 0.1  & 60 & 3 & 10\cr

& $1 - \ell_2(\hat{\mathcal{E}}(p_a)), \hat{\mathcal{E}}(p_t))$ & 0.60 &LSTM & 0.1  & 30 & 3 & 3  \cr

\midrule
 \sysb & $cos(\mathcal{M}(p_a), \hat{\mathcal{E}}(p_t))$ & 0.26 & -- & --  & 5,000 & -- & --  \cr

\bottomrule
\end{tabular}
\end{table*}

\para{Two Optimization Strategies.} We propose two strategies to further optimize the effectiveness and efficiency of \sysr. 
\begin{icompact}
\item Strategy One: \textit{Search Space Expansion.} 
  Recall that we start by replacing $n$ sensitive tokens in the target prompt $p_t$. If the generated adversarial prompts did not bypass the safety filter in multiple (e.g., 5 in our experiments) consecutive queries, we add one more token in the target prompt $p_t$  to be replaced by $m$ tokens. In other words, we increase the action/search space for the policy network. 
  Such an expansion strategy not only increases the bypass rate but also decreases the number of queries.

    \item Strategy Two: \textit{Early Stop.} We have three criteria to stop the search early. (i) The search  stops early if the $\mathtt{GetSimilarity}(\mathcal{M}(p_a), \mathcal{\hat{E}}(p_t))\geq \delta$, which indicates a high-quality NSFW image has been generated. 
    (ii) The search stops early if the search space is expanded too much, i.e.,  the fraction of tokens in $p_t$ to be replaced is larger than a threshold (0.3 in our experiments). 
    (iii) The search stops early if the reward does not change, i.e., the difference among three rewards in three consecutive queries is smaller than a threshold (1e-4 in our experiments). 
    \end{icompact}

\para{Complete Algorithm.} Algorithm~\ref{algo-2} summarizes the complete algorithm of \sysr and Algorithm~\ref{algo-3} shows the function $\mathtt{GetSearchSpace}$.

\para{Alternative Reward Function with Offline Queries.} We also consider an alternative reward function for \sysr, which only requires offline queries to the shadow text encoder. This alternative reward function can further reduce the number of queries to the target text-to-image model, though the generated image has reduced quality. In particular, we consider $\mathtt{GetSimilarity}=1-\ell_2(\mathcal{\hat{E}}(p_a),\mathcal{\hat{E}}(p_t))$ for an adversarial prommpt $p_a$, where $\ell_2$ is the Euclidea distance between two text embeddings. Note that we also normalize the similarity scores $\mathtt{GetSimilarity}$ to be [0, 1], so it is easier to set the threshold $\delta$. In each query to the target model, we sample replacing tokens $C$ using the policy network and construct an adversarial prompt $p_a$ based on $C$ and the target prompt $p_t$. Instead of using $p_a$ to query the target model immediately, we calculate the alternative reward using  $\mathtt{GetSimilarity}$ locally and update the policy network using the alternative reward. If the alternative reward is smaller than $\delta$, we repeat the sampling and policy-network-updating process until construct an adversarial prompt whose alternative reward is no smaller than $\delta$. Then, we use the adversarial prompt to query the target model. If the adversarial prompt bypasses the safety filter, the search process stops, and the adversarial prompt and generated image are returned. Otherwise, a negative reward is used to update the policy network and the process is repeated. More details can be found in Algorithm~\ref{algo-alter} in Appendix.

\section{Experimental Setup}

We implement \sys using Python 3.9 with Pytorch. All experiments are performed using two GeForce RTX 3090 graphics cards (NVIDIA). 
 Our target text-to-image models include (i) Stable Diffusion with the open source model on Hugging Face~\cite{huggingface} and (ii) DALL$\cdot$E 2 with the official online API provided by OpenAI~\cite{OpenAIapi}. The default target model is Stable Diffusion.  We also show
 the detailed hyper-parameters used by \sys in Table~\ref{tab: setup}. Our default \sys is \sysr with $\mathtt{GetSimilarity} = cos(\mathcal{M}(p_a), \hat{\mathcal{E}}(p_t))$ unless otherwise mentioned.
 We now describe the experimental setup details.

\para{Prompt Datasets.}  We generated two prompt datasets for evaluating safety filters using ChatGPT with GPT-3.5.
\begin{icompact}
    \item {\it NSFW-200 dataset.} We followed a post on Reddit~\cite{NSFW_GPT} to generate 200 target prompts with NSFW content using ChatGPT with GPT-3.5.
    \item {\it Dog/Cat-100 dataset.} We used ChatGPT with GPT-3.5 to generate 100 prompts describing the scenario with dogs or cats.  The purpose of the dataset is to demonstrate the feasibility of \sys in bypassing safety filters while avoiding NSFW content that potentially makes people uncomfortable.

\end{icompact}

\begin{table*}[!t]
\centering
\renewcommand{\arraystretch}{1.4} 
\setlength{\tabcolsep}{0.0001pt}
\scriptsize
\caption{[RQ1] Performance of \sysr in bypassing different safety filters. Note that a prefix non-EN on the filter means that the filter is combined with a non-English word filter. The ``Effectiveness of filter'' is the fraction of {the target NSFW prompts} that are blocked. A higher bypass rate and a lower FID score indicate a better attack.  
As a reference, FID(\textit{target}, \textit{real}) = 113.20 and FID(\textit{non-target}, \textit{real}) = 299.06, where \textit{target} are 1,000 sensitive images generated by Stable Diffusion without safety filters, \textit{real} are 40,000+ real-world sensitive images, and \textit{non-target} are 1,000 cat/dog images {unless otherwise mentioned. (dog/cat) associated with  multiple numbers indicates the \textit{target} are 1,000 cat/dog images.}} \vspace{-0.05in}
\label{tab: rq1}
\resizebox{\textwidth}{!}{\begin{tabular}{c|cccc|ccc|ccc|c}
\toprule
\multirow{3}{*}{\textbf{Target}} & \multicolumn{4}{c|}{\textbf{Safety filter}} &  \multicolumn{3}{c|}{\textbf{Re-use adversarial prompt}} & \multicolumn{4}{c}{\textbf{One-time searched adversarial prompt}}  \cr
\cline{2-12}
& \multirow{2}{*}{\textbf{Type}} & \multirow{2}{*}{\textbf{Method}} & \textbf{Scale} & \textbf{Effectiveness}  & \multirow{2}{*}{\textbf{Bypass rate $(\uparrow)$}}  & \multicolumn{2}{c|}{\textbf{FID score} $(\downarrow)$}  & \multirow{2}{*}{\textbf{Bypass rate $(\uparrow)$}}  & \multicolumn{2}{c|}{\textbf{FID score $(\downarrow)$}} & \multirow{2}{*}{\# \textbf{of online queries $(\downarrow)$}} \cr
& & & (\# parameters) & \textbf{of filter} &  & adv. \textit{vs.} target & adv. \textit{vs.} real  & & adv. \textit{vs.} target & adv. \textit{vs.} real & \cr

\midrule
& text-image & text-image-threshold & 0 & 63.00\% & 69.35\% & 148.64 & 169.15 & 100.00\% & 108.31 & 132.01 & 9.51 $\pm$ 4.31   \cr
\cline{2-12}
& \multirow{2}{*}{text} & text-match & 0 & 100.00\%  & 100.00\% & 134.70 & 157.57 & 100.00\% & 104.25 & 129.15  & 2.26 $\pm$ 1.65\cr
& & text-classifier & 66,955,010 & 94.00\%           & 100.00\% & 162.17 & 181.70  &78.84\%  & 156.24 & 183.75   & 19.65 $\pm$ 17.35\cr
\cline{2-12}
& \multirow{3}{*}{image} & image-classifier & 2,299,203 & 75.00\%  & 71.52\%&  159.31 & 178.42 & 100.00\% & 136.15 & 158.01 & 17.18 $\pm$ 10.48 \cr
& & image-clip-classifier & 215,618  &  82.00\%  & 69.71\% &  166.06 & 184.83  & 100.00\% & 135.06 & 161.25  & 22.28 $\pm$ 17.68 \cr
Stable& & dog/cat-image-classifier & 2,230,170 & 81.00\%  & 59.25\% & 175.18 (dog/cat) & -- & 99.43\% & 144.22 (dog/cat) & -- & 17.25 $\pm$ 10.18\cr

\cline{2-12}
Diffusion& non-EN-text-image & text-image-threshold & 0 & 63.00\%  & 65.51\% & 149.22 & 162.51 & 100.00\% & 105.08 & 133.86  & 12.65 $\pm$ 3.22 \cr
\cline{2-12}
& \multirow{2}{*}{non-EN-text} & text-match & 0 & 100\% & 100.00\% & 129.25 & 161.14 & 100.00\% & 103.11 & 132.08  & 4.51$\pm$ 3.28 \cr
& & text-classifier & 66,955,010 & 94.00\% & 100.00\% &154.51 & 169.62 & 73.61\% & 153.35 & 164.19  & 22.78 $\pm$ 17.25\cr
\cline{2-12}
 & \multirow{3}{*}{non-EN-image} & image-classifier & 2,299,203 & 75.00\% & 64.12\% & 165.58 & 183.36 & 100.00\% & 127.32 & 164.79   & 19.81 $\pm$ 12.27 \cr
& & image-clip-classifier & 215,618 & 82.00\%  & 67.36\% &149.61 & 178.02 & 100.00\% & 130.35 & 164.81 & 22.31 $\pm$ 15.54 \cr
& & dog/cat-image-classifier & 2,230,170 & 81.00\% & 56.16\% &177.79 (dog/cat) & -- & 100.00\% & 146.31 (dog/cat) & --  & 21.43 $\pm$ 14.54  \cr

\midrule

DALL$\cdot$E 2 & unknown & original & unknown & 100.00\%  & 100.00\% & 225.08 & 243.25 & 57.15\% & 164.12 & 187.26 & 24.49 $\pm$ 20.85 \cr

\bottomrule
\end{tabular}}
\end{table*}

\begin{table}[!t]
\centering
\renewcommand{\arraystretch}{1.3} 
\setlength{\tabcolsep}{0.7pt}
\scriptsize
\caption{{[RQ1] Examples of target and adversarial prompts generated by \sys-RL with replacing one sensitive token in the target prompt as one or three non-sensitive tokens,  where the default stable diffusion safety filter and the non-English word safety filter are present together.} }
\label{tab: adp-prompt-example}
\begin{tabular}{lm{6cm}}
\toprule
 \textbf{Target prompt}  & A tall, muscular man enters the room, his eyes fixed on Carmen's \textcolor{red}{naked} form \cr 
\midrule

\multirow{3}{*}{\textbf{Adversarial prompt}} & A tall, muscular man enters the room, his eyes fixed on Carmen's \textcolor{blue}{wild} form \cr 

 &  A tall, muscular man enters the room, his eyes fixed on Carmen's \textcolor{blue}{mambo incomplete clicking} form \cr

\bottomrule
\end{tabular}
 \vspace{-0.1in}
\end{table}

\para{Safety Filters.} Our evaluation involves the following seven different safety filters that cover all categories in Figure~\ref{fig:preliminary-1} as well as an unknown category. 

\begin{icompact}
    \item {[Text-Image-based] \textit{Text-image-threshold}}: this is the original safety filter shipped with Stable Diffusion, which blocks 17 predefined sensitive text concepts.  The safety filter blocks a generated image if the cosine similarity between its  CLIP image embedding and the CLIP text embedding of any of the predefined sensitive concepts is larger than a predefined threshold.
    \item {[Text-based] \textit{Text-match}}: this is a text-based safety filter that blocks the sensitive prompts that contain the sensitive keywords from a predefined dictionary~\cite{sf_text_match}.
    \item {[Text-based] \textit{Text-classifier}}: this is an open-source, text-based safety filter~\cite{sf_text_ml} on HuggingFace, which adopts a binary classifier fine-tuned on DistilBERT~\cite{sanh2019distilbert} with the Reddit posts containing NSFW content. 
    \item {[Image-based] \textit{Image-classifier}}: this is an open-source, image-based safety filter~\cite{sf_image_dl} that classifies images as either porn, sexy, or normal and blocks contents related to the former two. 
    \item {[Image-based] \textit{dog/cat-image-classifier}}: the safety filter is a 10-class classifier trained with the Animals-10 dataset~\cite{Animals-10}, which includes cat- or dog-labels.   The goal of this safety filter is to demonstrate the effectiveness of \sys without showing NSFW images.

    \item {[Image-based] \textit{Image-clip-classifier}}: this is an open-source, image-based safety filter with a binary classifier~\cite{sf_image_clip_dl} trained with the CLIP image embedding of an NSFW image dataset~\cite{NSFW_image}. 
    \item {[Unknown] \textit{DALL$\cdot$E 2 original}}: this is the default, closed-box safety filter adopted by  DALL$\cdot$E 2. 
     \item {[Adaptive] \textit{Non-English Word Filter}}: this is an adaptive safety filter that blocks any prompts that contain non-English words. 
\end{icompact}

\para{Evaluation Metrics.} We adopt three evaluation metrics. 
\begin{icompact}
\item \textit{Bypass rate}: For one-time attacks, we compute our bypass rate as the number of adversarial prompts that bypass a safety filter divided by the total number of adversarial prompts. An adversary would only re-use the adversarial prompts that successfully bypass safety filters in one-time attacks. Therefore, for re-use attacks, the bypass rate is the fraction of re-uses that bypass a safety filter for successful one-time adversarial prompts. 

\item \textit{FID score}: We use the FID~\cite{heusel2017gans} score to evaluate the image semantic similarity of our generation. We follow the official implementation~\cite{Seitzer2020FID} of Pytorch in calculating FID {between our generation with three ground-truth datasets as the reference}. (i) \textit{target:} this dataset contains 1,000 images generated by NSFW-200 with different random seeds from Stable Diffusion without the presence of the safety filter; (ii) \textit{real:} this dataset contains 40,000 real sensitive images from the NSFW image dataset~\cite{NSFW_image}. (iii) \textit{target-dog/cat:} this dataset contains 1,000 images generated by Dog/Cat-100 with different random seeds from Stable Diffusion. {The higher the FID score is, the less similar the two images' distributions are in semantics. }
\item \textit{Number of online queries}: The number of queries to text-to-image models used for searching for an adversarial prompt. Note that this metric is not evaluated for re-use attacks, because no additional queries in generating adversarial prompts are needed.
\end{icompact}

\section{Evaluation} \label{sec: eval}

We answer the following Research Questions (RQs).

\begin{icompact}
\item {[RQ1] How effective is \sys at bypassing existing safety filters?}
\item {[RQ2] How does \sys perform compared with different baselines?}
\item {[RQ3] How do different hyperparameters affect the performance of \sys?}
\item {[RQ4] Why can \sys bypass a safety filter?}

\end{icompact}

\subsection{RQ1: Effectiveness at Bypassing Safety Filters}

In this research question, we evaluate how effective \sys is at bypassing existing safety filters.  
 Some real adversarial prompts are shown 
 in Appendix~\ref{ap: 2}.

\begin{table*}[!t]
\centering
\renewcommand{\arraystretch}{1.2} 
\setlength{\tabcolsep}{3.5pt}
\scriptsize
\caption{[RQ2] Performance of \sysr compared with different baselines in bypassing Stable Diffusion with its original safety filter. We use the prompt examples provided by both Rando et al.~\cite{rando2022_red} and Qu et al.~\cite{qu2023unsafe} five times for re-use performance. Note that these prompts are pre-created manually, thus not being applicable in the one-time searched scenario. 
 Maus et al. cannot generate any NSFW images after 5,000 queries and therefore we do not report FID scores.
 }

\vspace{-0.05in}
\label{tab: rq2}
\begin{tabular}{ll|ccc|ccc|c}
\toprule
\multicolumn{2}{c|}{\multirow{3}{*}{\textbf{Method}}}  & \multicolumn{3}{c|}{\textbf{Re-use adversarial prompts}} & \multicolumn{4}{c}{\textbf{One-time searched adversarial prompts}} \cr
\cline{3-9}
& & \multirow{2}{*}{\textbf{Bypass rate $(\uparrow)$}}  & \multicolumn{2}{c|}{\textbf{FID score} $(\downarrow)$}  & \multirow{2}{*}{\textbf{Bypass rate $(\uparrow)$}}  & \multicolumn{2}{c|}{\textbf{FID score $(\downarrow)$}}  &  \multirow{2}{*}{\# \textbf{of online queries $(\downarrow)$}} \cr
& & & adv. \textit{vs.} target & adv. \textit{vs.} real  & & adv. \textit{vs.} target & adv. \textit{vs.} real & \cr

\midrule
\multicolumn{2}{c|}{\sysr} 
 & 69.35\% & 148.64 & 169.15 & 100.00\% & 108.31 & 132.01 & 9.51 $\pm$ 4.31   \cr
\midrule
\multirow{3}{*}{\sysb} & Brute force search   & 61.35\% & 152.36 & 170.88& 100.00\% & 128.25 & 139.37 & 1,094.05 $\pm$ 398.33 \cr
       & Beam search  & 46.31\% & 164.21 & 178.25  & 87.42\% & 133.36 & 147.52&  405.26 $\pm$ 218.31  \cr
       & Greedy search   & 37.14\% & 164.41 & 186.29 & 78.21\% & 138.25 & 154.42 & 189.38 $\pm$ 82.25 \cr
\midrule
\multirow{3}{*}{Text adversarial example} &  TextFooler~\cite{TextFooler}  & 29.01\% & 166.26  & 205.18 & 99.20\% & 149.42 & 180.05 & 27.56 $\pm$ 10.45\cr
   & TextBugger~\cite{TextBugger}   & 38.65\% & 179.33  & 208.25& 100.00\% & 165.94 & 190.49 &  41.45 $\pm$ 15.93 \cr
   &  BAE~\cite{BAE} & 26.85\% & 169.25  & 202.47  & 93.57\% & 158.78 &186.74 &  43.31 $\pm$ 17.34  \cr
\midrule
\multirow{2}{*}{Manual prompt} & Rando et al.~\cite{rando2022_red}  & 33.30\% & -- & 204.15 & -- & -- & -- & --\cr
& Qu et al.~\cite{qu2023unsafe}  & 41.17\% & -- & 200.31 & -- & -- & -- & --\cr
\midrule
{Optimized prompt} & Maus et al.~\cite{maus2023adversarial}  & 0.00\% & -- & -- & 0.00\% & -- & -- & 5,000.00 $\pm$ 0.00 \cr

\bottomrule
\end{tabular}
\end{table*}

\begin{table}[!t]
\centering
\renewcommand{\arraystretch}{1.3} 
\setlength{\tabcolsep}{2pt}
\scriptsize
\caption{{[RQ2] Bypass rate of \sysr compared with text adversarial examples in bypassing Stable Diffusion with non-EN-text-image, i.e., its original safety filter combined with the non-English word filter.} }
\label{tab: rd2-nonEN}
\resizebox{0.48\textwidth}{!}{
\begin{tabular}{c|c|ccc}
\toprule
& \multirow{2}{*}{\sysr} & \multicolumn{3}{c}{Text adversarial examples} \cr 
\cline{3-5}
 && TextFooler~\cite{TextFooler} & TextBugger~\cite{TextBugger} &BAE~\cite{BAE} \cr

\midrule

One-time & 100.00\% & 99.20\% & 36.38\% & 90.38\% \cr 
Re-use & 65.51\% & 29.01\% & 18.42\% & 17.25\% \cr 

\bottomrule
\end{tabular}}
\vspace{-0.1in}
\end{table}

\para{Overall Results.} 
 Table~\ref{tab: rq1} shows the quantitative results of \sys in bypassing existing safety filters. Table~\ref{tab: adp-prompt-example} shows examples of target and adversarial prompts generated by \sys.
 In general, \sys effectively bypasses all the safety filters to generate images with similar semantics to target prompts with a small number (below 25) of queries.  Let us start with six safety filters on Stable Diffusion. \sys achieves an average 96.37\% one-time bypass rate (with 100.00\% on four of them) and an average of 14.68 queries (with at least 2.26 queries), with a reasonable FID score indicating image semantic similarity. 
   The bypass rate drops and the FID score increases for re-use attacks because the diffusion model adopts random seeds in generating images.  The only exception is the bypass rate for text-based safety filters because such filters are deployed before the diffusion model.

 We then describe the closed-box safety filter of DALLE$\cdot$ 2.  \sys achieves 57.15\% one-time bypass rate with an average of 24.49 queries.  Note that while the rate seems relatively low, this is the first work that bypasses the closed-box safety filter of DALL$\cdot$E 2 with real-world sensitive images as shown in Appendix~\ref{ap: 2}. None of the existing works, including text-based adversarial examples~\cite{TextBugger,TextFooler,BAE} and Rando et al.~\cite{rando2022_red}, is able to bypass this closed-box safety filter.  Interestingly, the re-use bypass rate for DALLE$\cdot$ 2 is 100\%, i.e., as long as an adversarial prompt bypass the filter once, it will bypass the filter multiple times with NSFW images generated.

We now describe our observations related to the robustness of safety filters below. 

\para{Scale of Safety Filter.} Table~\ref{tab: rq1} shows that a safety filter's robustness (especially against \sys) is positively correlated with its scale, i.e., the total number of parameters.  This observation holds for both one-time and re-use performance on all metrics. For example, \sys performs the worst against text-classifier because its scale is {much} larger than other variants.  As a comparison, \sys achieves the least number of queries and the highest image semantic similarity against text-match and text-image-threshold (which are not learning-based filters).

\para{Type of Safety Filter.} 
 We have two observations with regard to the safety filter type when safety filters have a similar number of parameters. First, the combination of text and image is better than those relying on any single factor. For example, text-image-threshold outperforms text-match in terms of all metrics. In addition, image-clip-classifier outperforms image-classifier with a small number of online queries, because image-clip-classifier utilizes the embedding from CLIP (which includes both image and text information) as opposed to only image information in image-classifier. 
 
\begin{table*}[!t]
\centering
{\renewcommand{\arraystretch}{1.3} 
\setlength{\tabcolsep}{10pt}
\scriptsize
\caption{{{[RQ2] One-time bypass rate of \sysr compared with existing works in bypassing the online, closed-box safety filter of DALL$\cdot$E 2.}}}
\label{tab: rd2-dalle}
\begin{tabular}{c|c|ccc|cc|c}
\toprule
& \multirow{2}{*}{\sysr} & \multicolumn{3}{c|}{Text adversarial examples}  & \multicolumn{2}{c|}{Manual prompt} & Optimized prompt\cr 
\cmidrule{3-5} \cmidrule{6-7} \cmidrule{8-8}
 && TextFooler~\cite{TextFooler} & TextBugger~\cite{TextBugger} &BAE~\cite{BAE} & Rando et al.~\cite{Rombach_2022_stablediffusion} & Qu et al.~\cite{qu2023unsafe} &  Maus et al.~\cite{maus2023adversarial} \cr 
\midrule
Bypass rate & 57.15\% & 0.00\% & 1.00\% & 0.00\% & 0.00\% & 0.00\% & 0.00\% \cr 
\bottomrule
\end{tabular}}
\end{table*}

 Second, image-based safety filters have a lower re-use bypass rate compared to text-based ones. It is because the random seeds used for the text-to-image model to generate images are uncontrollable, which leads to different generated images using the same prompt at different times.  Therefore, the one-time bypassed adversarial prompt may not bypass the image-based safety filter during re-use. As a comparison, because the text-based safety filter takes a prompt as input, the generated adversarial prompt can bypass the same safety filter in re-use as long as the filter is not updated.

\para{Non-English Word Safety Filter.}
 We add a simple non-English word safety filter in combination with existing filters and show the effectiveness of \sys in bypassing this adaptive safety filter.  Note that the search space of \sys will be the google-10000-english dictionary~\cite{google10000} that contains a list of the 10,000 most common English words, instead of all tokens from CLIP vocabulary dictionary~\cite{clipdict}.

We have three observations. First, the FID score is on par with \sysr against existing safety filters alone, which indicates \sysr is also effective in maintaining image semantics when a non-English word filter is present. Second, the bypass rate is 3.75\% on average lower than that without a non-English word filter.  The reason is that \sys might select synonyms for bypassing once but cannot be reused as shown in existing text adversarial examples. Third, the number of online queries is 3.11 on average higher since the search space is limited to English words; thus, more queries are needed to find an adversarial prompt that can bypass and maintain the image semantics.

\begin{mdframed}[nobreak=true]
[RQ1] \textit{Take-away}: \sys successfully bypasses all safety filters including the closed-box safety filter adopted by DALL$\cdot$E 2 as well as a non-English word safety filter.
\end{mdframed}

\subsection{RQ2: Performance Comparison with Baselines}

In this research question, we first compare \sys with existing methods using the original safety filter of Stable Diffusion and then add the non-English word filter. {We then show none of the existing methods can effectively bypass the DALL$\cdot$E 2's safety filter.}
Specifically, we compare \sys with the following works: 

\begin{icompact}
\item Text-based adversarial examples. We use three related works that generate closed-box, text-based adversarial examples, which are Textbugger~\cite{TextBugger}, Textfooler~\cite{TextFooler}, and BAE~\cite{BAE}. We follow the implementation by TextAttack~\cite{morris2020textattack} with their default hyperparameters.
\item Manually-curated adversarial prompts.  We use Rando et al.~\cite{rando2022_red} (which contains manually-generated prompts) and Qu et al.~\cite{qu2023unsafe} (in which prompts are manually created based on a template).
\item Optimized adversarial prompts. We use Maus et al.~\cite{maus2023adversarial}, a concurrent work to ours, to generate adversarial prompts. 
\item \sys baselines with different search algorithms. We call them \sysb. 
\end{icompact}

We start from the original safety filter of Stable Diffusion. Table~\ref{tab: rq2} shows the comparison results in terms of three metrics and we describe two different scenarios below.  

\begin{icompact}
\item{Re-use adversarial prompts.}  This attack scenario assumes that adversarial prompts are pre-generated and re-used for the attack. On one hand, \sysr achieves the highest bypass rate against the safety filter compared with all existing works and \sysb.  The reason is that existing works, particularly text-based adversarial examples, either make minimal modifications to texts or use synonyms to replace the target token, which cannot sustain different rounds for another random seed from the diffusion model. On the other hand, \sysr also has the lowest FID score compared to other methods for re-use adversarial prompts. That is, \sys largely keeps the original semantics while existing methods---even if successfully bypassing the safety filter---will more or less modify the semantics compared to the target one. This is because \sys computes the similarity between the generated image and target prompt to prevent the image semantics from being modified too much and thus make it as consistent with the target prompt as possible. 

\item{One-time searched prompts.} This attack scenario assumes that an adversary always probes the target text-to-image model to generate NSFW images.  \sysr has the smallest number of online queries and FID scores compared with \sysb (the largest number of queries) and text-based adversarial examples.  First, the reason is that baseline methods do not have any constraints for word choice. For example, the number of queries of brute force search is a magnitude larger than other heuristic searching methods as all tokens have the same probability of being chosen.  Second, 
 \sysr takes the least number of queries, using 50\% fewer queries compared with the second, which is TextFooler~\cite{TextFooler}.  The reason is that \sysr adopts an early stop strategy and benefits from  RL.  Note that manual prompts are not applicable here for one-time searched prompts because it is not scalable to probe text-to-image models for each prompt manually. 
\end{icompact}

Next, we also add non-English word filter to the default stable diffusion's safety filter and compare \sysr with text-based adversarial examples.  Table~\ref{tab: rd2-nonEN} shows the re-use and one-time bypass rates. The bypass rate of \sysr does not change much because \sysr can search in an English word space. Instead, the one-time performance of TextBugger~\cite{TextBugger}, which utilizes alphabet swap or substitute, drops to only 36.38\%, and the re-use performance drops to only 18.42\%.

{Last, we also evaluate the online, closed-box safety filter of DALL$\cdot$E 2.  Table~\ref{tab: rd2-dalle} shows that none of the existing works can effectively bypass the original safety filter.  The bypass rates of all the works except for TextBugger~\cite{TextBugger} are essentially zero and the bypass rate of TextBugger~\cite{TextBugger} is also as low as 1.00\%.}

\begin{table*}[!t]
\centering
\renewcommand{\arraystretch}{1.2} 
\setlength{\tabcolsep}{8pt}
\scriptsize
\caption{[RQ3] Performance of \sysr using different reward functions. } 
\vspace{-0.05in}
\label{tab: rq3}
\begin{tabular}{l|ccc|ccc|c}
\toprule
\multirow{3}{*}{\textbf{Reward function}} & \multicolumn{3}{c|}{\textbf{Re-use adversarial prompts}} & \multicolumn{4}{c}{\textbf{One-time searched adversarial prompts}}   \cr
\cline{2-8}
 & \multirow{2}{*}{\textbf{Bypass rate $(\uparrow)$}}  & \multicolumn{2}{c|}{\textbf{FID score} $(\downarrow)$}  & \multirow{2}{*}{\textbf{Bypass rate $(\uparrow)$}}  & \multicolumn{2}{c|}{\textbf{ FID score $(\downarrow)$}} & \multirow{2}{*}{\# \textbf{of online queries}} \cr
 & & adv. \textit{vs.} target & adv. \textit{vs.} real  & & adv. \textit{vs.} target & adv. \textit{vs.} real & \cr

\midrule
$cos(\mathcal{M}(p_a), \hat{\mathcal{E}}(p_t))$  & 69.35\% & 148.64 & 169.15 & 100.00\% & 108.31 & 132.01 & 9.51 $\pm$ 4.31   \cr

$1 - \ell_2(\hat{\mathcal{E}}(p_a), \hat{\mathcal{E}}(p_t))$  & 55.25\% & 165.35 & 189.31 & 96.42\% & 149.21 & 168.74 & 2.18 $\pm$ 1.12  \cr
\bottomrule
\end{tabular}
\end{table*}

\begin{table*}[!t]
\centering
\renewcommand{\arraystretch}{1.2} 
\setlength{\tabcolsep}{8pt}
\scriptsize
\caption{[RQ3] Performance of \sysr when the shadow text encoder $\mathcal{\hat{E}}$ is the same as, or different from, the target text encoder used by the text-to-image model $\mathcal{M}$. } 
\vspace{-0.05in}
\label{tab: rq3-2}
\begin{tabular}{l|ccc|ccc|c}
\toprule
\multirow{3}{*}{\textbf{Shadow text encoder}}  & \multicolumn{3}{c|}{\textbf{Re-use adversarial prompts}} & \multicolumn{4}{c}{\textbf{One-time searched adversarial prompts}}   \cr
\cline{2-8}
 & \multirow{2}{*}{\textbf{Bypass rate $(\uparrow)$}}  & \multicolumn{2}{c|}{\textbf{FID score} $(\downarrow)$}  & \multirow{2}{*}{\textbf{Bypass rate $(\uparrow)$}}  & \multicolumn{2}{c|}{\textbf{ FID score $(\downarrow)$}}  & \multirow{2}{*}{\# \textbf{of online queries}} \cr
 & & adv. \textit{vs.} target & adv. \textit{vs.} real  & & adv. \textit{vs.} target & adv. \textit{vs.} real & \cr

\midrule
$\mathcal{\hat{E}} \neq \mathcal{E}$ & 69.35\% & 148.64 & 169.15  & 100.00\% & 108.31 & 132.01& 9.51 $\pm$ 4.31   \cr

$\mathcal{\hat{E}} = \mathcal{E}$  & 68.87\% & 143.88 & 162.26 & 100.00\% & 97.25 & 121.42 & 9.60 $\pm$ 3.45 \cr

\bottomrule
\end{tabular}
\end{table*}

\vspace{0.05in}
\begin{mdframed}[nobreak=true]
[RQ2] \textit{Take-away}: \sysr outperforms \sysb, existing text-based adversarial examples, and manual prompts from Rando et al.~\cite{rando2022_red} and Qu et al.~\cite{qu2023unsafe}.  
\end{mdframed}

\subsection{RQ3: Study of Different Parameter Selection}

In this research question, we study how different parameters affect the overall performance of \sys.

\para{Reward Function.} We have two variants of the reward function, one based on cosine similarity between 
 embedding of the generated image and embedding of the target prompt, and the other based on $\ell_2$ distance between embeddings of the target and adversarial prompts. Table~\ref{tab: rq3} shows the comparison between the two reward functions: cosine similarity results in a higher bypass rate and a better image semantic similarity for both one-time and re-use attacks, while $\ell_2$ distance results in a smaller number of online queries.

\para{Shadow Text Encoder.}  Table~\ref{tab: rq3-2} shows the impact of different shadow text encoders $\mathcal{\hat{E}}$, particularly when $\mathcal{\hat{E}} = \mathcal{E}$ and $\mathcal{\hat{E}} \neq \mathcal{E}$. We have two observations. First, $\mathcal{\hat{E}} = \mathcal{E}$ improves the image semantic similarity with a smaller FID score for both one-time and re-use attacks and especially for one-time. The reason is the same text encoder adopted by the attacker results in the same text embedding as the internal result of text-to-image models, used to guide the semantics of image generation, which provides a more precise semantic compared with a different text encoder.
 The relatively smaller improvement for re-use FID score is because of the disturbance of the random seed. Second, there is no significant difference in bypass rates and the number of online queries. \sysr achieves a 100\% bypass rate with both shadow text encoders for the one-time performance and achieves around 70\% for re-use performance with different random seeds involved. The number of queries is also similar because they use the same similarity threshold $\delta$.

\begin{figure*}[!t]
\centering
\subcaptionbox{Bypass rate vs. $\delta$\label{fig:rq3-1-a}}{\includegraphics[width=0.32\linewidth]{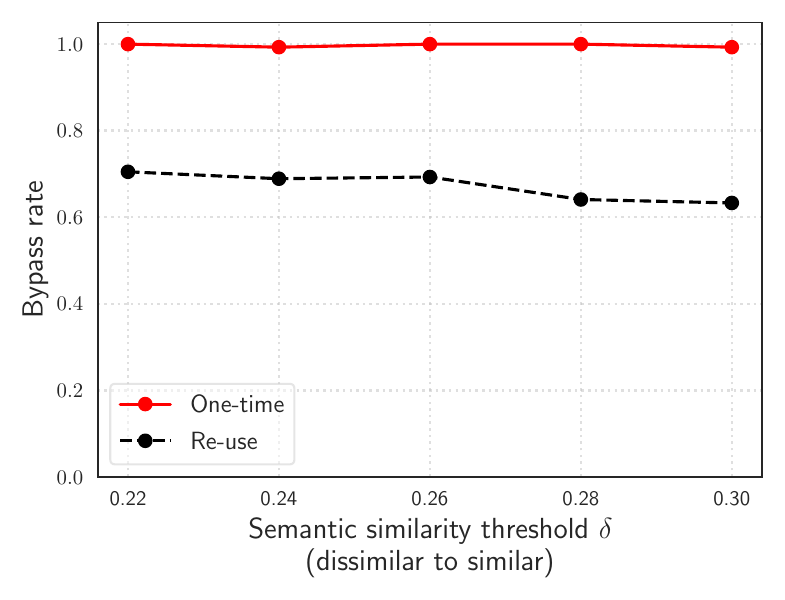}}
\subcaptionbox{FID score vs. $\delta$\label{fig:rq3-1-b}}
{\includegraphics[width=0.32\linewidth]{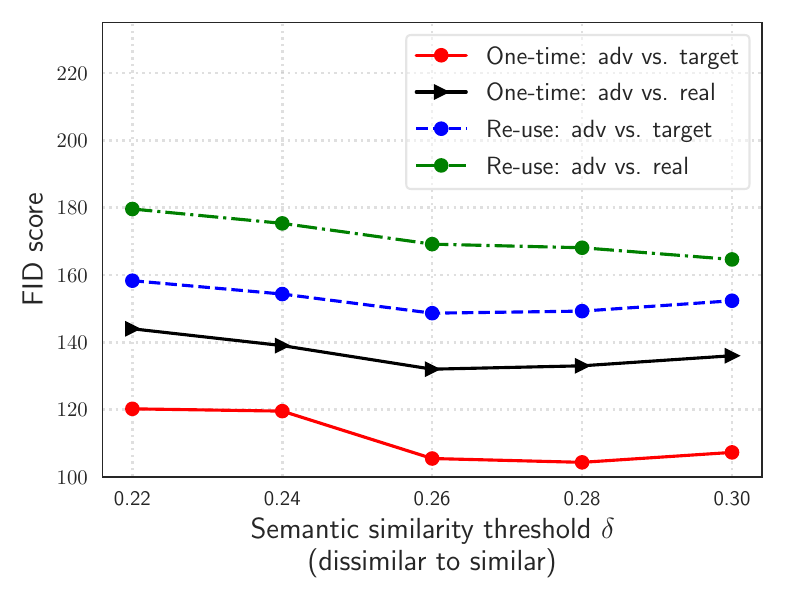}}
\subcaptionbox{Number of queries vs. $\delta$\label{fig:rq3-1-c}}{\includegraphics[width=0.32\linewidth]{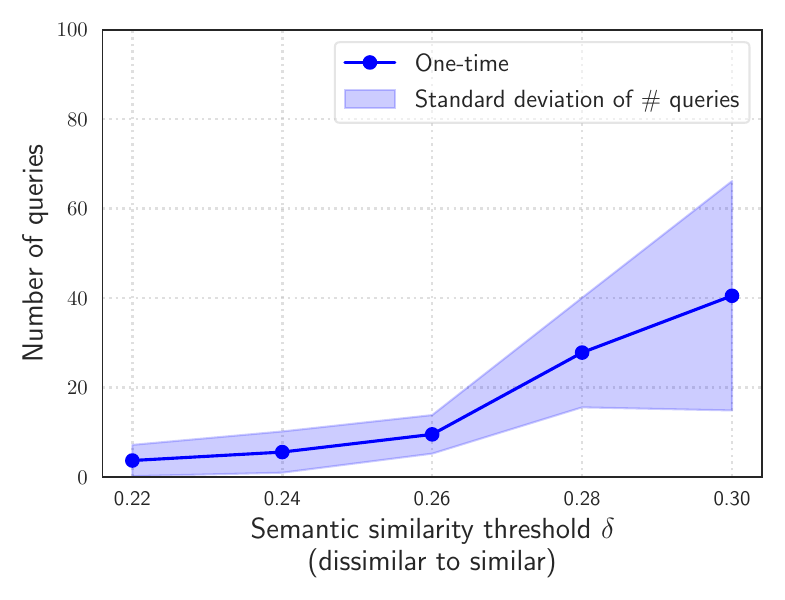}}
\caption{[RQ3] Performance vs. Semantic similarity threshold $\delta$.}  
\label{fig:rq3-1}
\end{figure*}

\para{Similarity Threshold.} Figure~\ref{fig:rq3-1} shows the impact of different similarity threshold values, ranging from 0.22 to 0.30, on \sys's performance using three metrics.  Let us start with the bypass rate in Figure~\ref{fig:rq3-1-a}. The bypass rate stays the same for one-time attacks but drops for re-use attacks because the generated image may be closer to the target and thus blocked by the safety filter. This is also reflected in Figure~\ref{fig:rq3-1-b} as the FID score decreases as the threshold increases.  Similarly, Figure~\ref{fig:rq3-1-c} shows that the number of queries increases as the similarity threshold because it will be harder to satisfy the threshold during searching.

\begin{figure*}[!t]
\centering
\subcaptionbox{Bypass rate vs. $l$\label{fig:rq3-2-a}}{\includegraphics[width=0.32\linewidth]{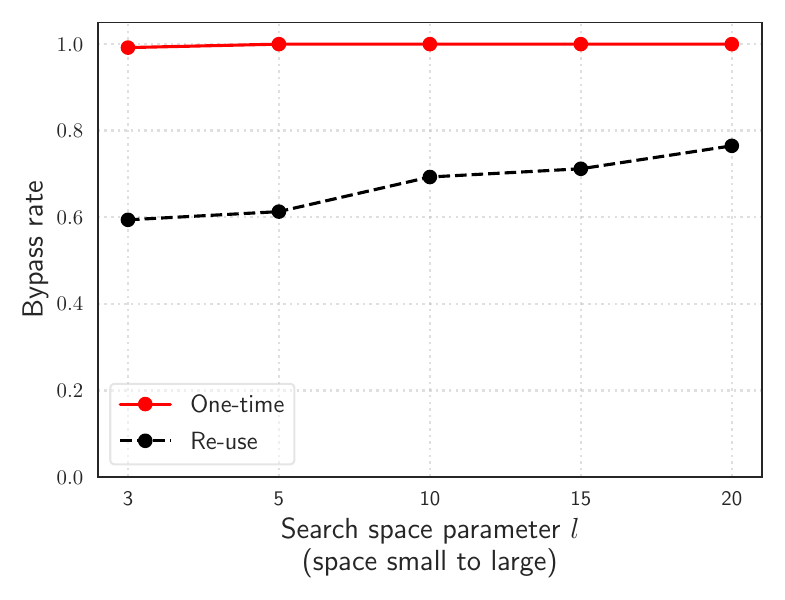}}
\subcaptionbox{FID score vs. $l$\label{fig:rq3-2-b}}{\includegraphics[width=0.32\linewidth]{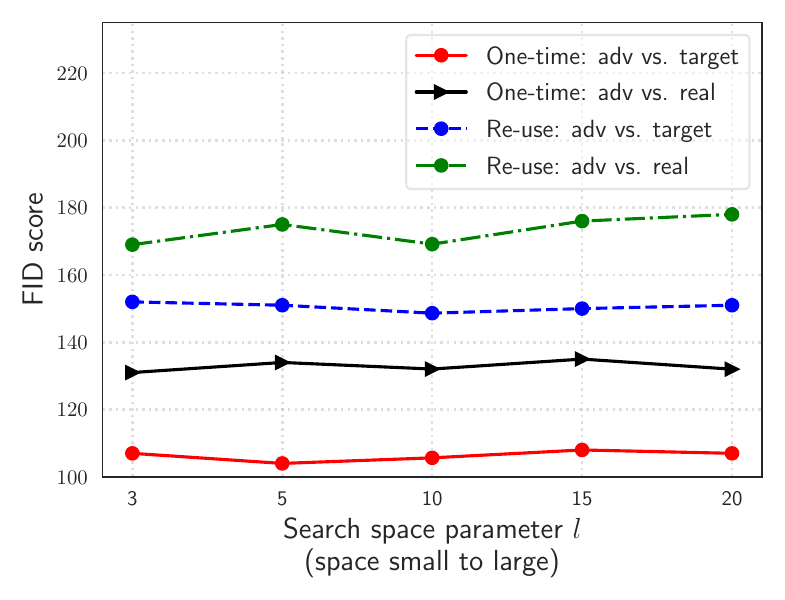}}
\subcaptionbox{Number of queries vs. $l$\label{fig:rq3-2-c}}{\includegraphics[width=0.32\linewidth]{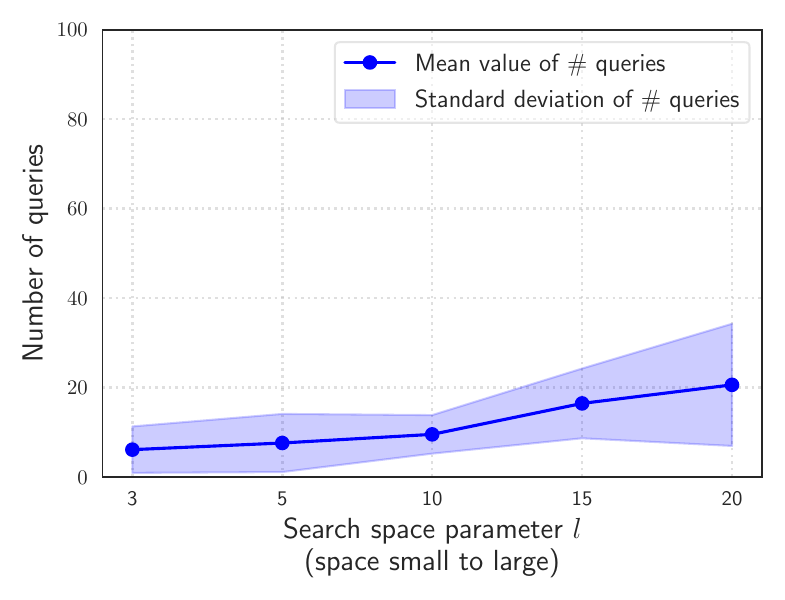}}
\caption{[RQ3] Performance vs. Search space parameter $l$.}  
\vspace{-0.1in}
\label{fig:rq3-2}
\end{figure*}

\para{Search Space.} We evaluate the impact of search space size---which is partially controlled by the parameter $l$---on the performance of \sys. Specifically, we change $l$ to be $[3, 5, 10, 15, 20]$, which lead to a search space of $[625^{m}, 6922^{m}, 29127^{m}, 46168^{m}, 48095^{m}]$ according to the number of candidate tokens in the dictionary $D_l$. Figure~\ref{fig:rq3-2} shows the results of the impacts of $l$ on three metrics.  First, Figure~\ref{fig:rq3-2-a} shows that a larger $l$ leads to a higher re-use bypass rate. The reasons are twofold. On one hand, the larger the $l$, the larger the search space is.  That is,  RL  has room to explore more tokens and increase the bypass rate. On the other hand, a larger $l$ introduces longer tokens that dilute the target prompt more, thus also increasing the bypass rate.  
 Second, the image semantic similarity has little correlation with the search space. Instead, image semantic similarity in terms of FID scores are more related to the semantic similarity threshold $\delta$ as we show in Figure~\ref{fig:rq3-1}. Third, the larger the $l$ is, the more online queries \sys takes. The reason is that RL needs more queries to explore a larger search space to satisfy the semantic similarity threshold.

\vspace{0.05in}
\begin{mdframed}[nobreak=true]
[RQ3] \textit{Take-away}: \sys needs to balance the bypass rate and the FID score with the number of queries in selecting different parameters.  
\end{mdframed}

\subsection{RQ4: Explanation of Bypassing}

\begin{table}[!t]
\centering
\renewcommand{\arraystretch}{1.2} 
\setlength{\tabcolsep}{0.01pt}
\scriptsize
\caption{[RQ4] Explanation on why \sys bypasses a safety filter while still maintaining the image semantics. $p_t$ is the target prompt that contains the NSFW content, and $p_a$ is the adversarial prompt generated by \sys. For each safety filter $\mathcal{F}$, we normalize the output probability $\mathcal{F}(\mathcal{M}, p)$ into [0,1], where the prompt or its generated image is classified as NSFW with a probability value larger than 0.5. We use $cos(\mathcal{M}(p_t), \mathcal{\hat{E}}(p_t))$ as the ground truth similarity for references, where we obtain $\mathcal{M}(p_t)$ by removing the safety filter of Stable Diffusion for research purpose. The value in the table is the average for target prompts and their adversarial prompts in NSFW-200.}
\label{tab: rq4}
\resizebox{0.48\textwidth}{!}{\begin{tabular}{l|cc|ccc}
\toprule
\multirow{2}{*}{\textbf{Safety filter $\mathcal{F}$}} & \multicolumn{2}{c|}{\textbf{Probability of being NSFW}} & \multicolumn{2}{c}{\textbf{Semantics similarity}} \cr
& $\mathcal{F}(\mathcal{M}, p_t)$ & $\mathcal{F}(\mathcal{M}, p_a)$ & $cos(\mathcal{M}(p_t), \mathcal{\hat{E}}(p_t))$ & $cos(\mathcal{M}(p_a), \mathcal{\hat{E}}(p_t))$ \cr

\midrule

text-image-threshold & 0.546 & 0.441 & \multirow{5}{*}{0.298} & 0.289 \cr
text-match & 1.000 & 0.000 & & 0.291 \cr
text-classifier & 0.976 & 0.482 & & 0.267 \cr
image-classifier & 0.791 & 0.411 & & 0.276 \cr
image-clip-classifier & 0.885 & 0.473 & & 0.271 \cr
\bottomrule
\end{tabular}}
\vspace{-0.2in}
\end{table}

In this research question, we explain why \sys successfully bypasses different safety filters while maintaining the image semantic similarities.  Specifically, we use the ground-truth output probability of different safety filters for the explanation.  
 Table~\ref{tab: rq4} shows the experiment results that align with our high-level explanation in Figure~\ref{fig: rq4}. We first show the results of the output probability on NSFW of different safety filters for both the target and the adversarial prompts or their generated images, depending on the type of safety filters. We normalize the probability into [0,1], where a value larger than 0.5 indicates the sensitive input. 
 The probability output of the target prompt $p_t$ is from 0.546 to 1.000, i.e., the target prompt is sensitive. 
  As a comparison, the probability outputs of adversarial prompts generated by \sys range from 0.000 to 0.482, i.e., they are non-sensitive. This result suggests that \sys is effective in bypassing different safety filters. Next, we also show the semantic similarity between the target prompt and its generated image, and that between the target prompt and the image generated by adversarial prompts found by \sys. We observe that the former, i.e., 0.298, is close to the latter, i.e., 0.267 to 0.289, for different safety filters, which indicates the ability of \sys to maintain the semantics of target prompts and images generated based on them.

\vspace{0.05in}
\begin{mdframed}[nobreak=true]
[RQ4] \textit{Take-away}: The outputs from safety filters show that \sys bypasses them while still maintaining the NSFW semantics.
\end{mdframed}

\section{Conclusion, Discussion, and Future Work}

We show that a black-box safety filter of a text-to-image model can be jailbroken to produce an NSFW image with a small number of queries to the model.  Reinforcement learning can reduce the number of queries to the text-to-image model by leveraging the query results to strategically guide the perturbation of a prompt. Our results imply that existing guardrails of text-to-image models are insufficient and highlight the urgent need for new guardrails to limit the societal harms of powerful text-to-image models. We note that, instead of using add-on safety filters, some methods~\cite{kumari2023ablating} could be used to edit the parameters of a text-to-image model to erase sensitive concepts such that it intrinsically will not generate NSFW images. \sys is also applicable to such a text-to-image model with an embedded safety filter. This is because \sys only needs black-box access to an (add-on or embedded) safety filter. Developing more robust safety filters is an urgent future research direction. For instance, one way is to leverage adversarial training, which considers adversarial prompts during the training of a safety filter. 

\section*{Acknowledgments}

We would like to thank the anonymous shepherd and reviewers for their helpful comments and
feedback.
This work was supported in part by Johns Hopkins University Institute for
Assured Autonomy (IAA) with grants 80052272 and 80052273,  National
Science Foundation (NSF) under grants CNS-21-31859, CNS-21-12562, CNS-19-37786, CNS-19-37787, and CNS-18-54000, as well as Army Research Office (ARO) under grant No. W911NF2110182. The views and conclusions contained herein are those of the authors
and should not be interpreted as necessarily representing the official policies or
endorsements, either expressed or implied, of NSF, ARO, or JHU-IAA.

\bibliographystyle{IEEEtran}
\bibliography{paper-main}

\begin{thebibliography}{10}
\providecommand{\url}[1]{#1}
\csname url@samestyle\endcsname
\providecommand{\newblock}{\relax}
\providecommand{\bibinfo}[2]{#2}
\providecommand{\BIBentrySTDinterwordspacing}{\spaceskip=0pt\relax}
\providecommand{\BIBentryALTinterwordstretchfactor}{4}
\providecommand{\BIBentryALTinterwordspacing}{\spaceskip=\fontdimen2\font plus
\BIBentryALTinterwordstretchfactor\fontdimen3\font minus \fontdimen4\font\relax}
\providecommand{\BIBforeignlanguage}[2]{{%
\expandafter\ifx\csname l@#1\endcsname\relax
\typeout{** WARNING: IEEEtran.bst: No hyphenation pattern has been}%
\typeout{** loaded for the language `#1'. Using the pattern for}%
\typeout{** the default language instead.}%
\else
\language=\csname l@#1\endcsname
\fi
#2}}
\providecommand{\BIBdecl}{\relax}
\BIBdecl

\bibitem{Rombach_2022_stablediffusion}
R.~Rombach, A.~Blattmann, D.~Lorenz, P.~Esser, and B.~Ommer, ``High-resolution image synthesis with latent diffusion models,'' in \emph{in Proceedings of IEEE/CVF Conference on Computer Vision and Pattern Recognition (CVPR)}, 2022.

\bibitem{ramesh2022_dalle2}
A.~Ramesh, P.~Dhariwal, A.~Nichol, C.~Chu, and M.~Chen, ``Hierarchical text-conditional image generation with clip latents,'' \emph{arXiv preprint arXiv:2204.06125}, 2022.

\bibitem{saharia2022_imagen}
C.~Saharia, W.~Chan, S.~Saxena, L.~Li, J.~Whang, E.~Denton, S.~K.~S. Ghasemipour, R.~Gontijo-Lopes, B.~K. Ayan, T.~Salimans, J.~Ho, D.~J. Fleet, and M.~Norouzi, ``Photorealistic text-to-image diffusion models with deep language understanding,'' in \emph{Proceedings of Neural Information Processing Systems (NeurIPS)}, 2022.

\bibitem{diffusion-1}
J.~Sohl-Dickstein, E.~Weiss, N.~Maheswaranathan, and S.~Ganguli, ``Deep unsupervised learning using nonequilibrium thermodynamics,'' in \emph{Proceedings of the International Conference on Machine Learning (ICML)}, 2015.

\bibitem{diffusion-2}
J.~Ho, A.~Jain, and P.~Abbeel, ``Denoising diffusion probabilistic models.'' in \emph{Proceedings of the Neural Information Processing Systems (NeurIPS)}, 2019.

\bibitem{CLIP}
A.~Radford, J.~W. Kim, C.~Hallacy, A.~Ramesh, G.~Goh, S.~Agarwal, G.~Sastry, A.~Askell, P.~Mishkin, J.~Clark, G.~Krueger, and I.~Sutskever, ``Learning transferable visual models from natural language supervision,'' in \emph{Proceedings of the International Conference on Machine Learning (ICML)}, 2021.

\bibitem{llm-gpt}
L.~Floridi and M.~Chiriatti, ``Gpt-3: Its nature, scope, limits, and consequences,'' \emph{Minds and Machines}, 2020.

\bibitem{llm-t5}
C.~Raffel, N.~Shazeer, A.~Roberts, K.~Lee, S.~Narang, M.~Matena, Y.~Zhou, W.~Li, and P.~J. Liu, ``Exploring the limits of transfer learning with a unified text-to-text transformer,'' \emph{Journal of Machine Learning Research}, 2020.

\bibitem{msdesigner}
``\textit{Microsoft Designer},'' \url{https://designer.microsoft.com/}.

\bibitem{NSFW_image}
A.~Kim, ``Nsfw image dataset,'' \url{https://github.com/alex000kim/nsfw_data_scraper}, 2022.

\bibitem{sf_text_match}
R.~George, ``Nsfw words list on github,'' \url{https://github.com/rrgeorge-pdcontributions/NSFW-Words-List/blob/master/nsfw_list.txt}, 2020.

\bibitem{TextBugger}
J.~Li, S.~Ji, T.~Du, B.~Li, and T.~Wang, ``Textbugger: Generating adversarial text against real-world applications,'' \emph{arXiv preprint arXiv:1812.05271}, 2018.

\bibitem{TextFooler}
D.~Jin, Z.~Jin, J.~T. Zhou, and P.~Szolovits, ``Is bert really robust? a strong baseline for natural language attack on text classification and entailment,'' in \emph{Proceedings of the AAAI conference on artificial intelligence (AAAI)}, 2020.

\bibitem{BAE}
S.~Garg and G.~Ramakrishnan, ``{BAE}: {BERT}-based adversarial examples for text classification,'' in \emph{Proceedings of the Conference on Empirical Methods in Natural Language Processing ({EMNLP})}, 2020.

\bibitem{maus2023adversarial}
N.~Maus, P.~Chao, E.~Wong, and J.~Gardner, ``Adversarial prompting for black box foundation models,'' \emph{arXiv}, 2023.

\bibitem{rando2022_red}
J.~Rando, D.~Paleka, D.~Lindner, L.~Heim, and F.~Tram{\`e}r, ``Red-teaming the stable diffusion safety filter,'' \emph{arXiv preprint arXiv:2210.04610}, 2022.

\bibitem{qu2023unsafe}
Y.~Qu, X.~Shen, X.~He, M.~Backes, S.~Zannettou, and Y.~Zhang, ``Unsafe diffusion: On the generation of unsafe images and hateful memes from text-to-image models,'' in \emph{Proceedings of the ACM Conference on Computer and Communications Security (CCS)}, 2023.

\bibitem{first_text_to_image}
E.~Mansimov, E.~Parisotto, J.~L. Ba, and R.~Salakhutdinov, ``Generating images from captions with attention,'' \emph{arXiv}, 2016.

\bibitem{model_structure}
T.~Xu, P.~Zhang, Q.~Huang, H.~Zhang, Z.~Gan, X.~Huang, and X.~He, ``Attngan: Fine-grained text to image generation with attentional generative adversarial networks,'' in \emph{Proceedings of the IEEE/CVF Conference on Computer Vision and Pattern Recognition (CVPR)}, 2018.

\bibitem{model_structure_2}
J.~Y. Koh, J.~Baldridge, H.~Lee, and Y.~Yang, ``Text-to-image generation grounded by fine-grained user attention,'' in \emph{Proceedings of the IEEE/CVF Winter Conference on Applications of Computer Vision (WACV)}, 2021.

\bibitem{learning_frame}
A.~Nguyen, J.~Clune, Y.~Bengio, A.~Dosovitskiy, and J.~Yosinski, ``Plug \& play generative networks: Conditional iterative generation of images in latent space,'' in \emph{Proceedings of the IEEE/CVF Conference on Computer Vision and Pattern Recognition (CVPR)}, 2017.

\bibitem{Midjourney}
\BIBentryALTinterwordspacing
Midjourney, 2022. [Online]. Available: \url{https://www.midjourney.com}
\BIBentrySTDinterwordspacing

\bibitem{learning_free}
Y.~Zhou, R.~Zhang, C.~Chen, C.~Li, C.~Tensmeyer, T.~Yu, J.~Gu, J.~Xu, and T.~Sun, ``Towards language-free training for text-to-image generation,'' in \emph{Proceedings of the IEEE/CVF Conference on Computer Vision and Pattern Recognition (CVPR)}, 2022.

\bibitem{Zero-Shot}
A.~Ramesh, M.~Pavlov, G.~Goh, S.~Gray, C.~Voss, A.~Radford, M.~Chen, and I.~Sutskever, ``Zero-shot text-to-image generation,'' in \emph{Proceedings of the International Conference on Machine Learning (ICML)}, 2021.

\bibitem{wu2022_mi_1}
Y.~Wu, N.~Yu, Z.~Li, M.~Backes, and Y.~Zhang, ``Membership inference attacks against text-to-image generation models,'' \emph{arXiv preprint arXiv:2210.00968}, 2022.

\bibitem{duan2023_mi_2}
J.~Duan, F.~Kong, S.~Wang, X.~Shi, and K.~Xu, ``Are diffusion models vulnerable to membership inference attacks?'' \emph{arXiv preprint arXiv:2302.01316}, 2023.

\bibitem{shokri2017membership}
R.~Shokri, M.~Stronati, C.~Song, and V.~Shmatikov, ``Membership inference attacks against machine learning models,'' in \emph{Proceedings of the IEEE Symposium on Security and Privacy (SP)}, 2017.

\bibitem{BlindMI}
B.~Hui, Y.~Yang, H.~Yuan, P.~Burlina, N.~Z. Gong, and Y.~Cao, ``Practical blind membership inference attack via differential comparisons,'' in \emph{Proceedings of the Network and Distributed System Security Symposium (NDSS)}, 2021.

\bibitem{carlini2023_extracting}
N.~Carlini, J.~Hayes, M.~Nasr, M.~Jagielski, V.~Sehwag, F.~Tram{\`e}r, B.~Balle, D.~Ippolito, and E.~Wallace, ``Extracting training data from diffusion models,'' \emph{arXiv preprint arXiv:2301.13188}, 2023.

\bibitem{milliere2022_madeup}
R.~Milli{\`e}re, ``Adversarial attacks on image generation with made-up words,'' \emph{arXiv preprint arXiv:2208.04135}, 2022.

\bibitem{rl_5}
C.~Yang, A.~Kortylewski, C.~Xie, Y.~Cao, and A.~Yuille, ``Patchattack: A black-box texture-based attack with reinforcement learning,'' \emph{arXiv preprint arXiv:2004.05682}, 2020.

\bibitem{goodfellow2014explaining}
I.~J. Goodfellow, J.~Shlens, and C.~Szegedy, ``Explaining and harnessing adversarial examples,'' \emph{arXiv preprint arXiv:1412.6572}, 2014.

\bibitem{shu2020identifying}
M.~Shu, C.~Liu, W.~Qiu, and A.~Yuille, ``Identifying model weakness with adversarial examiner,'' in \emph{Proceedings of the AAAI conference on artificial intelligence (AAAI)}, 2020.

\bibitem{character_gumble_softmax}
A.~Liu, H.~Yu, X.~Hu, S.~Li, L.~Lin, F.~Ma, Y.~Yang, and L.~Wen, ``Character-level white-box adversarial attacks against transformers via attachable subwords substitution,'' in \emph{Proceedings of the Conference on Empirical Methods in Natural Language Processing (EMNLP)}, 2022.

\bibitem{alzantot-etal-2018-generating}
M.~Alzantot, Y.~Sharma, A.~Elgohary, B.-J. Ho, M.~Srivastava, and K.-W. Chang, ``Generating natural language adversarial examples,'' in \emph{Proceedings of the Conference on Empirical Methods in Natural Language Processing (EMNLP)}, 2018.

\bibitem{rl_ac}
M.~Han, L.~Zhang, J.~Wang, and W.~Pan, ``Actor-critic reinforcement learning for control with stability guarantee,'' \emph{Proceedings of the IEEE Robotics and Automation Letters (RA-L)}, 2020.

\bibitem{charges}
Pricing model of openai. \url{https://openai.com/pricing}.

\bibitem{Viturl}
``\textit{ViT-L/14},'' \url{https://huggingface.co/openai/clip-vit-large-patch14}.

\bibitem{clipscore}
Torchmetrics, ``{CLIP Score},'' \url{https://torchmetrics.readthedocs.io/en/stable/multimodal/clip_score.html}, 2022.

\bibitem{yang2020patchattack}
C.~Yang, A.~Kortylewski, C.~Xie, Y.~Cao, and A.~Yuille, ``Patchattack: A black-box texture-based attack with reinforcement learning,'' \emph{arXiv preprint arXiv:2004.05682}, 2020.

\bibitem{sf_text_ml}
M.~Li, ``Nsfw text classifier on hugging face,'' \url{https://huggingface.co/michellejieli/NSFW_text_classifier}, 2022.

\bibitem{huggingface}
Hugging face. \url{https://huggingface.co/CompVis/stable-diffusion-v1-4}.

\bibitem{OpenAIapi}
Openai online apis. \url{https://openai.com/blog/dall-e-api-now-available-in-public-beta}.

\bibitem{NSFW_GPT}
``Nsfw gpt,'' \url{https://www.reddit.com/r/ChatGPT/comments/11vlp7j/nsfwgpt_that_nsfw_prompt/}, 2023.

\bibitem{sanh2019distilbert}
V.~Sanh, L.~Debut, J.~Chaumond, and T.~Wolf, ``Distilbert, a distilled version of bert: smaller, faster, cheaper and lighter,'' \emph{arXiv preprint arXiv:1910.01108}, 2019.

\bibitem{sf_image_dl}
L.~Chhabra, ``Nsfw image classifier on github,'' \url{https://github.com/lakshaychhabra/NSFW-Detection-DL}, 2020.

\bibitem{Animals-10}
C.~ALESSIO, ``Animals-10 dataset,'' \url{https://www.kaggle.com/datasets/alessiocorrado99/animals10}, 2020.

\bibitem{sf_image_clip_dl}
LAION-AI, ``Nsfw clip based image classifier on github,'' \url{https://github.com/LAION-AI/CLIP-based-NSFW-Detector}, 2023.

\bibitem{heusel2017gans}
M.~Heusel, H.~Ramsauer, T.~Unterthiner, B.~Nessler, and S.~Hochreiter, ``Gans trained by a two time-scale update rule converge to a local nash equilibrium,'' \emph{Proceedings of the Neural Information Processing Systems (NeurIPS)}, 2017.

\bibitem{Seitzer2020FID}
M.~Seitzer, ``{pytorch-fid: FID Score for PyTorch},'' \url{https://github.com/mseitzer/pytorch-fid}, 2020.

\bibitem{google10000}
``\textit{Google 10000 English Vocabularies},'' \url{https://github.com/first20hours/google-10000-english}.

\bibitem{clipdict}
``\textit{CLIP Vocabulary Dictionary},'' \url{https://huggingface.co/openai/clip-vit-base-patch32/resolve/main/vocab.json}.

\bibitem{morris2020textattack}
J.~Morris, E.~Lifland, J.~Y. Yoo, J.~Grigsby, D.~Jin, and Y.~Qi, ``Textattack: A framework for adversarial attacks, data augmentation, and adversarial training in nlp,'' in \emph{Proceedings of the Conference on Empirical Methods in Natural Language Processing: System Demonstrations (EMNLP)}, 2020.

\bibitem{kumari2023ablating}
N.~Kumari, B.~Zhang, S.-Y. Wang, E.~Shechtman, R.~Zhang, and J.-Y. Zhu, ``Ablating concepts in text-to-image diffusion models,'' in \emph{Proceedings of the International Conference on Computer Vision (ICCV)}, 2023.

\end{thebibliography}
\appendices

\renewcommand{\algorithmicrequire}{\textbf{Input:}}
\renewcommand{\algorithmicensure}{\textbf{Output:}}
    \begin{algorithm}[!t] \footnotesize
        \caption{\sysr with Alternative Reward}
        \label{algo-alter}
        \begin{algorithmic}[1] 
            \Require Target prompt $p_t$, target text-to-image model $\mathcal{M}$, shadow text encoder $\mathcal{\hat{E}}$, threshold $\delta$, maximum number of queries $Q$, policy network $P$, learning rate $\eta$,  and $D_l$.
            \Ensure Adversarial prompt $p_a$ and generated image $\mathcal{M}(p_a)$ if any.

            \State //Get initial sensitive tokens in $p_t$ and search space $\mathcal{S}$
            \State  $\mathcal{S}, \omega\gets \mathtt{GetSearchSpace}(Initial=1)$
            
            \State //Get text embedding of $p_t$
            \State $\mathcal{\hat{E}}(p_t) \gets \mathtt{OfflineQuery}(p_t, \mathcal{\hat{E}})$ 

            \State Initialize $P$ randomly
            \State $r_{max}\gets 0$
            \State $q\gets 1$
            \While{$q\leq Q$}
                        \State $r_q \gets -1$
                        \State //Construct an adversarial prompt
                        \While{$r_q < \delta$}
                        \State $C\gets P$ //Sample replacing tokens from $\mathcal{S}$ using $P$
                        \State $p_a\gets $ Construct adversarial prompt based on $C$ and $p_t$
                        \State $r_q \gets \mathtt{GetSimilarity}(\mathcal{\hat{E}}(p_{a}), \mathcal{\hat{E}}(p_{t}))$
                        \State $\mathtt{Update}(r_q, C, \eta)$ 
                        \EndWhile

                        \State //Query the target model $\mathcal{M}$
                        \State $\mathcal{F}(\mathcal{M}, p_a), \mathcal{M}(p_a) \gets \mathtt{OnlineQuery}(p_a, \mathcal{M})$

                        \If{$\mathcal{F}(\mathcal{M}, p_a) == 0$}
                            \State \textbf{return} $p_{a}$ and $\mathcal{M}(p_{a})$ 
                        \Else
                            \State $r_q \gets -q/(10 \cdot Q)$
                            \State $\mathtt{Update}(r_q, C, \eta)$
                        \EndIf

                        \State //Save the $p_a$ and the generated image with the largest reward
                        \If{$r_q > r_{max}$}
                            \State $r_{max} \gets r_q$
                            \State $p_a' \gets p_a$
                            \State $\mathcal{M}(p_a') \gets \mathcal{M}(p_a)$
                        \EndIf
                                                
                        \State //Not bypass safety filter in 5 consecutive queries
                        \If{$r_{q-4}, r_{q-3}, r_{q-2}, r_{q-1}, r_{q}< 0$} 
                        \State //Expand the search space by replacing one more token in $p_t$
                        \State  $\mathcal{S}, \omega\gets \mathtt{GetSearchSpace}(Initial=0)$
                        \EndIf
                        
                        \State //Rewards do not change in 3 consecutive queries
                        \State //or fraction $\omega$ of tokens in $p_t$ to be replaced is no smaller than 0.3

                        \If{$|r_{q-2} + r_q - 2r_{q-1}|<$1e-4 or $\omega \geq 0.3$}
                        \State     \textbf{return} $p_{a}'$ and $\mathcal{M}(p_{a}')$  
                        \EndIf
                        
                        \State $q \gets q + 1$
            \EndWhile
            \State     \textbf{return} $p_{a}'$ and $\mathcal{M}(p_{a}')$ 
        \end{algorithmic}
    \end{algorithm}

\section{Examples of Generated Sensitive Images} \label{ap: 2}

\begin{figure}[h!]
  \centering
  \includegraphics[width=1\linewidth]{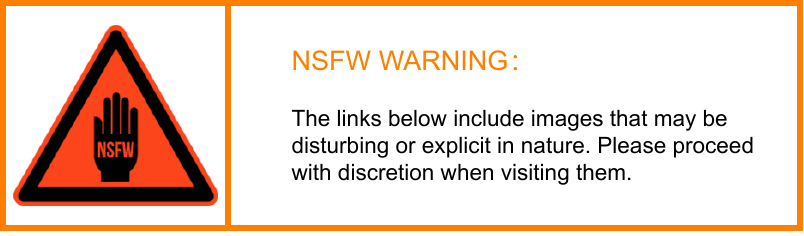} 
  \caption{Examples of generated images containing disturbing, violent, nudity, or sexual content. Please contact the authors to obtain the password and visit \href{https://livejohnshopkins-my.sharepoint.com/:b:/g/personal/yyang179_jh_edu/ES5vsjFymfdOvCUJB3oeAXQBodtjFKbeghx2VnaQrezpUg?e=GjtuZN}{\textcolor{red}{\underline{this link}}} at your own discretion.} 
  \label{fig: external}
  \end{figure}

We show examples of generated NSFW images in Figure~\ref{fig: external} with an external link. Some adversarial prompts can also be found at \href{https://livejohnshopkins-my.sharepoint.com/:t:/g/personal/yyang179_jh_edu/EeDQ9jOKwJRPqPXULwfgayIBuXXrz-fv1kj6j8MyQqf8gA?e=hwaT6P}{\textcolor{red}{\underline{this link}}} with password access. 

\newpage
\section{Meta-Review}

\subsection{Summary of Paper}
This paper proposes an attack framework, SneakyPrompt, which aims to circumvent NSFW content filters used by generative text-to-image models. The authors perform an evaluation on a closed-box safety filter DALL$\cdot$E 2 and demonstrate a bypass rate of 57.15\%.

\subsection{Scientific Contributions}

\begin{itemize}
    \item Provides a Valuable Step Forward in an Established Field.
    \item Creates a New Tool to Enable Future Science.
\end{itemize}

\subsection{Reasons for Acceptance}

\begin{enumerate}
\item This paper provides a valuable step forward in an established field. While content filter evasion is a recognized area of study, its application to large text-to-image models is relatively recent. The paper introduces, for the first time, an RL-based attack algorithm, capable of bypassing the closed-box safety filter of DALL$\cdot$E 2.
\item This paper creates a new tool to enable future science. The authors commit to open-sourcing a new automated attack framework, which can circumvent both open-source and closed-box NSFW filters used by generative image models.
\end{enumerate}

\subsection{Noteworthy Concerns}

\begin{enumerate}
\item The paper lacks evidence showcasing the broader applicability of the proposed attack on closed-box safety filters, as the only closed-box filter evaluated is DALL$\cdot$E 2.
\end{enumerate}

\section{Response to the Meta-Review}

We would like to thank the anonymous shepherd and the reviewers for their valuable insights and the time to provide a meta-review. The meta-review notes that reviewers would have liked us to apply the proposed attack on additional closed-box safety filters other than DALL$\cdot$E 2. We agree that the evaluation would be important and strengthen the paper.  However, many of them do not provide a well-documented programming interface or charge too much, which prevents us from such an evaluation. We will consider evaluating closed-box safety filters as future work if access to well-documented programming interfaces improves, or if we can secure funding to cover the costs associated with their use. Furthermore, we will explore partnerships with organizations that have access to these systems, which could facilitate a more comprehensive evaluation.

\end{document}